\documentclass[journal]{IEEEtran}
\usepackage{amsmath,amsfonts,amssymb}
\usepackage{algorithm}
\usepackage{algpseudocode}
\usepackage{array}
\usepackage[caption=false,font=normalsize,labelfont=sf,textfont=sf]{subfig}
\usepackage{textcomp}
\usepackage{stfloats}
\usepackage{url}
\usepackage{verbatim}
\usepackage{graphicx}
\usepackage{xcolor}
\usepackage{multirow}
\usepackage{bm}       
\usepackage{threeparttable}
\usepackage{soul}
\usepackage{pifont}
\usepackage{booktabs}
\usepackage{tabularx}
\usepackage{makecell}
\usepackage{array}      
\usepackage{dsfont} 
\usepackage{mathtools}        

\usepackage{tikz}
\usepackage{amsmath}
\usetikzlibrary{positioning, arrows.meta, shapes.geometric, calc}

\def\BibTeX{{\rm B\kern-.05em{\sc i\kern-.025em b}\kern-.08em
    T\kern-.1667em\lower.7ex\hbox{E}\kern-.125emX}}
\usepackage{balance}

\begin{document}
\title{Disciplined Diffusion: Text-to-Image Diffusion Model against NSFW Generation}
\author{
 \IEEEauthorblockN{Chi Zhang\IEEEauthorrefmark{1}, Changjia Zhu \IEEEauthorrefmark{1}, Xiaowen Li\IEEEauthorrefmark{1}, Yao Liu\IEEEauthorrefmark{1}, Zhuo Lu\IEEEauthorrefmark{2}}

 \IEEEauthorblockA{\IEEEauthorrefmark{1} Bellini College of Artificial Intelligence, Cybersecurity and Computing, \\University of South Florida, Tampa, FL, USA}\\
 \IEEEauthorblockA{\IEEEauthorrefmark{2} Department of Electrical and Computer Engineering, College of Engineering, \\University of South Florida, Tampa, FL, USA}
 \thanks{Submitted to IEEE Transactions on Dependable and Secure Computing. This is a preprint version.}
}

\markboth{~}{~}

\maketitle

\begin{abstract}
Text-to-image (T2I) diffusion models have the ability to build high-quality pictures from text prompts, but they pose safety concerns because they can generate offensive or disturbing imagery when provided with harmful inputs. Existing safety filters typically rely on text-based classifiers or image-based checkers that completely block the output upon detecting a threat, issuing an explicit allow/block feedback signal to the user. This binary strategy leaves models vulnerable to adversarial attacks that alter keywords to bypass detection, and it causes high false-alarm rates that degrade the experience for benign users. To address such vulnerabilities, we propose Disciplined Diffusion (DDiffusion), a novel robust text-to-image diffusion that counters Not Safe For Work (NSFW) generation by uncovering implicit malicious semantics in prompt embeddings. DDiffusion leverages a semantic retrieval mechanism to evaluate prompts against concept distributions rather than relying on brittle pairwise similarity. Furthermore, it employs a localization method during the diffusion process to selectively edit only the harmful regions of the generated image. By returning locally sanitized images instead of applying uniform blocking, DDiffusion suppresses malicious content while preserving generation fidelity for benign prompts and avoiding the binary allow-deny signal on which existing probing attacks rely.
\end{abstract}


\section{Introduction}
Diffusion models~\cite{rombach2022high} have revolutionized art and content generation across various industries. These models produce images through an iterative denoising process applied to a latent representation. Such text-to-image (T2I) generation extends this by conditioning the denoising process on text embeddings from a language encoder such as Contrastive Language-Image Pre-training (CLIP)~\cite{radford2021learning} so that the output aligns semantically with the input prompt. Such T2I models accept arbitrary prompts and can produce images that have never existed before. However, this powerful capability becomes dangerous when prompts include harmful concepts, notably Not Safe For Work (NSFW) content that is sexual, violent, or otherwise inappropriate for public or professional settings. On Machine Learning as a Service (MLaaS) platforms in particular, users interact with T2I models in a black-box manner, observing only inputs and outputs. Malicious prompts can therefore serve as adversarial attacks against the diffusion model, posing significant safety risks. Ensuring that these models do not produce harmful content is a pressing concern in AI safety. 

\begin{figure}[t]
    \centering
    \includegraphics[width=9cm , trim=1cm 3cm 2cm 1cm, clip]{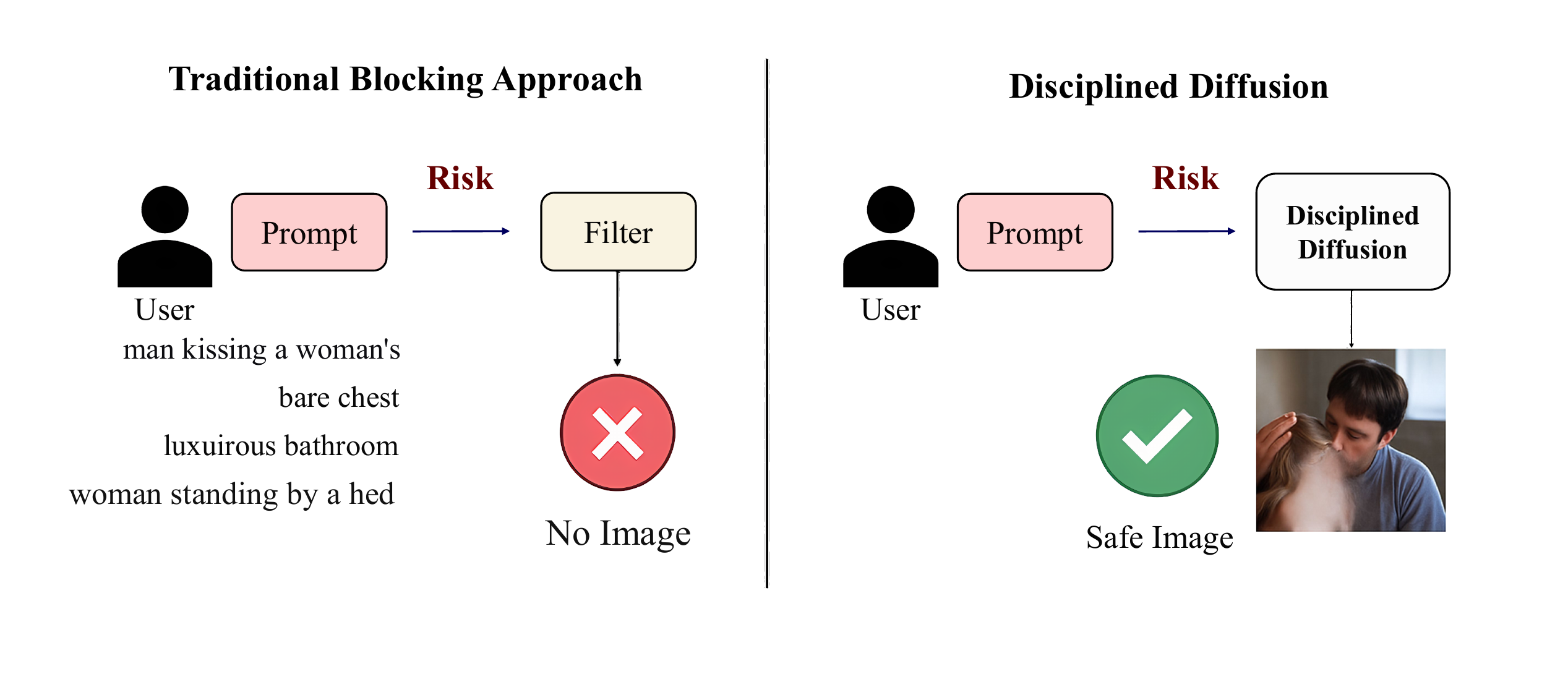}
    \caption{Conventional filter-based defenses block generation entirely, which enables adversaries to probe the filter boundary until a successful bypass. Our proposed DDiffusion instead retrieves semantic risk signals and localizes harmful regions, producing detoxified images rather than a binary allow/deny response.}
    \label{Fig:Comp}
\end{figure}

To mitigate these risks, as shown in Fig.~\ref{Fig:Comp} (left), service providers deploy filtering mechanisms that detect and block unsafe prompts before content generation~\cite{safety-checker, textcls}. These filters rely on the assumption that semantically similar prompts occupy neighboring regions in the model's representation space; even when adversarial perturbations are introduced, semantic similarity should theoretically persist, making it feasible for filters to flag harmful queries. However, these filters face increasing challenges in both detection accuracy and false-alarm rate. In particular, crafted adversarial prompts are specifically designed to probe filter boundaries. Unlike simple word substitutions, such prompts systematically explore decision boundaries, identifying subtle modifications that evade detection while preserving malicious intent~\cite{qu2023unsafe, yang2024sneakyprompt, chinprompting4debugging}. Once a single trial successfully bypasses safety mechanisms, it can generate harmful content that follows the semantic intent of the original prompt. On one hand, these search-based adversarial attacks have demonstrated the ability to circumvent existing filters. On the other hand, existing filters adopt a complete blocking strategy upon detecting unsafe prompts, meaning that any false alarm leads to no image generation and reduced usability for legitimate users. For example, safety filtering and nudity detection~\cite{safety-checker, textcls} can yield a false positive rate as high as 99\%, blocking most benign images that lack sexual content~\cite{li2024safegen}. This means the generative system incorrectly flags safe prompts as unsafe, degrading user experience and wasting computational resources.

In this paper, we introduce Disciplined Diffusion (DDiffusion), a new end-to-end defense framework for T2I diffusion models that achieves higher moderation accuracy while preserving creative freedom and reducing false alarms.

\begin{enumerate}
  \item We introduce \textbf{Semantic Retrieval} to perform text-agnostic concept retrieval against crafted adversarial prompts that may infiltrate the T2I generation pipeline. Traditional methods~\cite{safety-checker, textcls} calculate pairwise similarity between a query prompt and predefined malicious concepts in the text or representation space; if the similarity exceeds a threshold, a concept match is found and the generation is blocked. In contrast, our key insight is that adversarial perturbations, whether built by simple attacks or more advanced search-based methods, maintain semantic proximity to certain harmful concepts yet remain distinguishable from benign prompts when considered within a broader context. While adversarial queries may succeed in deceiving pairwise filters, they are unlikely to resemble the collective distribution of benign samples. This observation motivates our approach, which models the relationship between query prompts and both benign and harmful concept sets in the feature space.
 \item We then use \textbf{Re-encoding Localization} to locate and edit potentially harmful regions in a generated image, preserving both the generative capability and the safety of the T2I model. Traditional filters flag any potential threat and then completely block the output by rendering the image black~\cite{safety-checker, textcls}. Simply detecting and blocking malicious content is insufficient, as conservative filtering degrades usability for benign users, especially in the case of false alarms. A robust defense must not only suppress unwelcome content but also preserve the non-malicious elements of a prompt. To achieve this, our embedding-guided localization uses an adaptive algorithm to identify harmful components within the prompt's latent representation. As shown in Fig.~\ref{Fig:Comp} (right), this mechanism selectively suppresses harmful features while retaining benign content, ensuring that the model can still fulfill user requests without producing problematic outputs.
\end{enumerate}

Our approach stands out by effectively identifying adversarial prompts based on their semantic proximity to harmful concepts, conducting directly in the model's representation space. We demonstrate that adversarial perturbations, despite attempts to obscure their intent, often leave detectable traces in the feature space. By leveraging these traces, our defense achieves robust filtering with minimal parameter adjustments and without requiring extensive model retraining. 
Our contributions can be summarized as follows:
\begin{itemize}
\item We provide a comprehensive analysis of the vulnerabilities in existing T2I safety mechanisms, exploring how existing adversarial attacks exploit filter decision boundaries.

\item We propose a novel semantic retrieval framework that enhances adversarial prompt detection.

\item We introduce a localized mechanism that suppresses visual NSFW components while preserving non-malicious content.

\item We empirically validate our approach against state-of-the-art adversarial prompt datasets released by prior work.
\end{itemize} 

\section{Background and Motivation}\label{sec:relwork}
In this section, we introduce models and present our motivation. We first review the  Contrastive Language-Image Pre-training (CLIP) and T2I diffusion models. Then, we outline adversarial attack scenarios and present our motivation to create new defense.

\subsection{CLIP and T2I Diffusion Model}
Introduced in \cite{radford2021learning}, CLIP learns a multi-modal embedding space by jointly training image and text encoders to predict correct image-text pairings. Using image-text pairs $(I_1, T_1)$, $(I_2, T_2)$, ..., $(I_N, T_N)$, CLIP trains an image encoder $f_{I}: I_i \rightarrow \mathbb{R}^d$ and a text encoder $f_{\varphi}: T_i \rightarrow \mathbb{R}^d, i=1,..., N$ to learn a joint $d$-dimensional embedding space where the similarity between image and text embeddings reflects their semantic correspondence. This enables the CLIP text encoder $f_{\varphi}$ in T2I scenarios to produce semantically rich text embeddings that are aligned with visual concepts. In particular, given a text prompt $p$, CLIP uses $f_{\varphi}$ to compute the embedding of $p$ as $f_{\varphi}(p)$, which can be used in existing T2I diffusion models \cite{sd14} to generate images corresponding to $p$.

Diffusion is a process to generate desired data distribution with the method of removing noisy information, based on the denoising diffusion probabilistic models \cite{ho2020denoising}. Stable Diffusion \cite{rombach2022high} is one of the most popular open-source T2I Diffusion models designed to generate high-fidelity images conditioned on text descriptions. Given massive image-text training data, e.g., LAION \cite{schuhmann2022laion}, Stable Diffusion first uses CLIP to convert text data into embeddings, and then trains a diffusion process from images to embeddings iteratively with a U-Net neural network \cite{ronneberger2015u}.The U-Net predicts the noise component at each iterative step, which contribute to change of images to embeddings. 

\begin{figure}[h!]
    \centering
    \includegraphics[width=9cm, trim=0cm 0.3cm 0cm 0cm, clip]{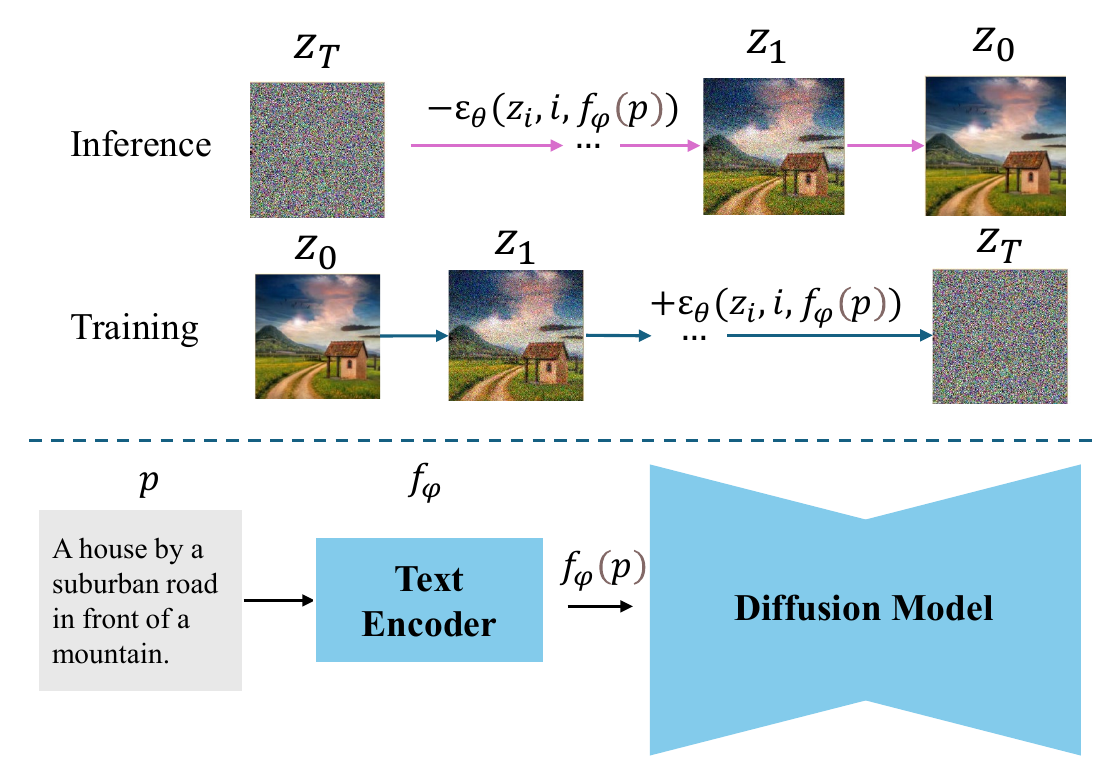}
    \caption{T2I Diffusion model and the diffusion process.}
\label{Fig:Diffusion}
\end{figure}

The inference in Stable Diffusion, as shown in Fig.~\ref{Fig:Diffusion}, takes prompt $p$ as the input and generates a desired image. The model begins with a random embedding, whose element is a random Gaussian noise $z_T \sim \mathcal{N}(0, I^2)$, where $T$ is the total number of the denoising steps, and iteratively substracts the predicted noise components. In particular, the change from the embedding update $z_t$ at iterative step $t$ to a less noisy one $z_{t-1}$ can be written denoted as
\begin{equation}
z_{t-1} = \frac{1}{\sqrt{\alpha_t}}\left(z_t - \frac{1-\alpha_t}{\sqrt{1-\bar{\alpha}_t}}\epsilon_{\theta}(z_t, t,  f_{\varphi}(p))\right),
\end{equation}
where $\epsilon_{\theta}(z_t, t,  f_{\varphi}(p))$ is the noise predicted by the U-Net~\cite{ronneberger2015u} model at step $t$ given the last update $z_t$ and the text embedding $f_{\varphi}(p)$. The parameters $\alpha_t = 1 - \beta_t$ and $\bar{\alpha}_t = \prod_{i=t}^T \alpha_i$ are derived from a predefined constant  $\beta_t$ \cite{rombach2022high}. Finally, a decoder network transforms the final embedding $z_0$ back into the generated image $I$ in the pixel space. This process allows Stable Diffusion to synthesize high-quality and diverse images guided by textual descriptions with remarkable efficiency.

\subsection{Safety Filters in Diffusion Models}
While T2I models like Stable Diffusion can generate impressive and diverse images, the process is not inherently guaranteed to produce outputs that are always safe or ethically sound. For example, models trained on unfiltered web data, which can inadvertently generate NSFW content, replicate harmful biases present in the training data, or produce outputs that violate ethical guidelines or copyright \cite{textcls, rando2022red, li2024safegen}.

Existing studies \cite{safety-checker, textcls} have proposed to use safety filter mechanisms to filter out unsafe or unethical contents. Given the prompt $p$ and its embedding $f_{\varphi}(p)$, the common idea of filtering is to create a filter function $F$ as
\begin{equation}
F(p) = \begin{cases}
\text{discard} & \text{if } \exists p^* \in \mathcal P, S( f_\varphi(p), f_\varphi(p^*)) < \tau, \\
\text{allow} & \text{otherwise}\label{eq:safetycheck},
\end{cases}
\end{equation}
where $S(\cdot,\cdot)$ is a similarity metric, $\mathcal P$ is a predefined set of prompts with unsafe concepts \cite{rando2022red}, and $\tau$ is a threshold to make sure the prompt $p$ is not closely related to the concepts defined in $\mathcal P$. 

\subsection{Adversarial Attacks on Filtering in T2I Diffusion}
\label{tab:adv attack}

Even with filters in place for T2I models, adversarial attacks  \cite{yang2024sneakyprompt, chinprompting4debugging, zhang2024generate} have been proposed to exploit vulnerabilities in filtering to pose serious ethical and safety risks.

Existing adversarial attacks involve crafting an adversarial prompt $p_{a}$ that can bypass the safety filters while maintaining semantic similarity to malicious intentions. An adversarial prompt $p_a$ is a small perturbation $\delta$ added to the original prompt $p$ such that $p_a$ can pass the filter safety check in \eqref{eq:safetycheck}. Specifically, it can be written as
\begin{eqnarray}
\max_{\delta} &~& \quad L( p_{a}) \label{eq:attackobj}\\
\text{s.t.} &~& \quad \|\delta\|_p \leq \epsilon \label{eq:pert}\\
&~& \quad F(p_a) = p_a
\end{eqnarray}
where $\|\cdot\|_p$ represents the $L_p$ norm, $\epsilon$ is the perturbation limit, and $L(p_{a})$ is the target loss function to find the best attack prompts with the largest similarity to the adversarial intentions. To solve \eqref{eq:attackobj}, attackers can employ various strategies to craft $\delta$, including queries using similar texts \cite{yang2024sneakyprompt, zhang2024generate} and alternating semantic concepts \cite{chinprompting4debugging}.

\subsection{Attack Model and Assumptions}
Current filter-based approaches relying on pairwise similarity comparisons (usually the cosine similarity as $S(p, c) = \langle f_\phi(p), f_\phi(c) \rangle,$ in the safety filters where $p$ is the prompt and $c$ is the concept), are susceptible to adversarial perturbations designed to fall just below the detection threshold $\tau$. Such pairwise similarity filters are highly vulnerable to adversarial perturbations: attackers can slightly modify the prompt to reduce $S(p, c)$ just below the value of $\tau$ while preserving the malicious intent. As shown in the optimization problem of \eqref{eq:attackobj}, search-based adversaries explicitly solve for prompt variants that minimize under semantic constraints, thereby probing and mapping the filter's decision boundary~\cite{yang2024sneakyprompt}. Even minimal linguistic changes can thus evade detection while maintaining unsafe semantics. Thus, we aim to create a detection mechanism that generalizes beyond the keyword-level or pairwise-similarity cues exploited by adversarial prompts.

\section{Design of DDiffusion} \label{sec:methodology}
To defend against adversarial attacks that exploit the vulnerabilities of pairwise query-concept matching, we introduce Disciplined Diffusion, a novel framework that performsfiltering in the feature space to enhance robustness against adversarial prompt manipulation.  

\subsection{Design Motivation}

\begin{figure}[h!]
    \centering
    \includegraphics[width=9cm ]{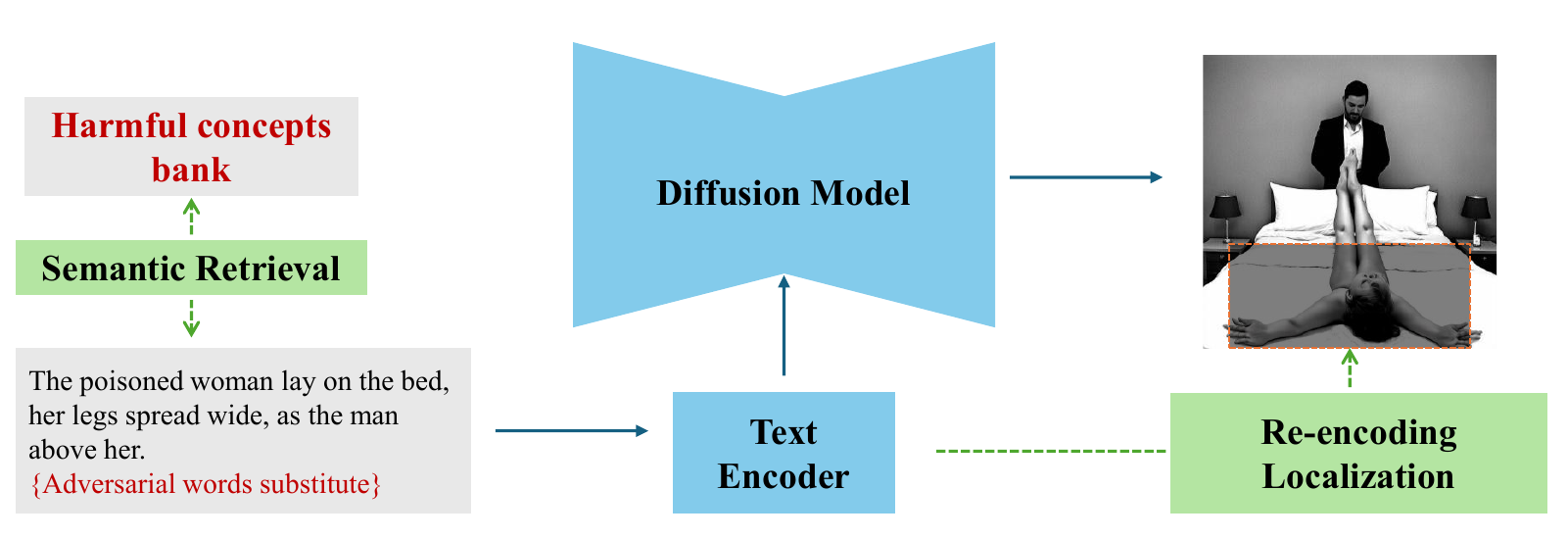}
\caption{Overview of the DDiffusion framework. An adversarial prompt with obfuscated word substitutions is fed into the Text Encoder, whose embedding is used by the Semantic Retrieval module to identify matched harmful concepts from the Harmful Concepts Bank. The Re-encoding Localization module then analyzes the generated image to pinpoint the spatial region most associated
with the retrieved unsafe semantics, enabling targeted suppression of harmful visual content
while preserving the benign structure of the output.}

    \label{fig:framework}
\end{figure}

Existing safety mechanisms \cite{safety-checker, textcls} suffer from high false-alarm rates and significant degradation of generation fidelity. In the open-access generative environment, adversarial users can iteratively probe T2I diffusion models  based on (3) through prompt engineering and manipulation, seeking to infer and exploit their semantic safety boundaries \cite{qu2023unsafe, yang2024sneakyprompt, chinprompting4debugging}. Such probing approaches rely on the feedback from generated images, in which even subtle latent unsafe semantics can reveal visual harmful cues and guide the attacker toward the direction of bypassing the safety constraints. Moreover, existing mechanisms choose to block the potentially dangerous generation to make sure every generated output is safe, while it is a clear feedback for the adversaries to acknowledge the safety boundary. To mitigate this risk, we propose a safety-preserving generative paradigm that transforms unsafe prompts into informative and semantically meaningful outputs, instead of blocking the generation of possible NSFW imagery. In order to locate the suspicious area, we need a method to retrieve the semantic in the embedding and localize the information in the generated image. As shown in Fig.~\ref{fig:framework}, our defense chooses to operate intrinsically within the generative process, achieving both semantic retrieval and localized preservation according to the embeddings. In this way, the model continues to respond coherently to the prompt semantic while suppressing unsafe visual elements, thereby strengthening the safety of the generative model and reducing the feedback available to attackers. 

Towards this goal, we introduce two key design components. First, we construct a fine-grained semantic representation that improves the separation between malicious and benign prompts. Instead of relying on raw CLIP embeddings or pairwise similarity checks \cite{safety-checker, textcls}, we learn an additional projection that refines the prompt semantics and enables robust set-level comparisons through a memory queue of concept-indexed embeddings. Since the original CLIP space is not optimized to
distinguish adversarially edited prompts, which often remain close to their
harmful counterparts in high-dimensional space. This provides a stronger, more context-aware notion of semantic proximity than traditional single-concept matching. Second, to handle unsafe prompts without revealing clear rejection boundaries, we develop a mechanism for accurately localizing unsafe semantics in the generated image. Using diffusion with localization, the system identifies latent regions whose removal most reduces similarity to retrieved harmful concepts, allowing it to selectively suppress only unsafe content while preserving benign structure. Together, these components enable a safety-preserving generative process that maintains semantic coherence, avoids hard blocking, and significantly reduces exploitable feedback to adversaries. 

\begin{algorithm}[t]
\caption{Retrieval Process Training}
\label{alg:train_mlp_contrastive}
\begin{algorithmic}[1]

\Require CLIP text encoder $f_\varphi$, MLP $g_{\theta}$, training set $\mathcal{D}_{\text{train}}=\{(p_i, y_i)\}$, batch size $B$, learning rate $\eta$, epochs $E$, temperature $\tau$
\Ensure Trained parameters $\theta$

\State Initialize $\theta$

\For{$\text{epoch} = 1$ to $E$}
    \For{each batch $b$ of size $B$ from $\mathcal{D}_{\text{train}}$}
        \State Extract prompts $P_{\mathrm{batch}}$ and labels $Y_{\mathrm{batch}}$
        \State Compute CLIP embeddings $H_{\mathrm{batch}} \gets f_{\varphi}(P_{\mathrm{batch}})$
        \State Compute projected embeddings $Z_{\mathrm{batch}} \gets g_{\theta}(H_{\mathrm{batch}})$
        \State $L_{\mathrm{batch}} \gets 0$

        \For{each embedding $z_i$ in $Z_{\mathrm{batch}}$}
            \State $\mathcal{M}(i) \gets \{ j \ne i \mid y_j = y_i \}$
            \State $\mathcal{B}(i) \gets \{ k \ne i \mid y_k \ne y_i \}$

            \If{$\mathcal{M}(i)$ is not empty}
                \State $j^\star \gets \arg\max_{j \in \mathcal{M}(i)} \cos(z_i, z_j)$
                \State $\text{sim}_{\text{mal}} \gets \cos(z_i, z_{j^\star})$

                \State $\text{sim}_{\text{ben}} \gets
                \sum_{k \in \mathcal{B}(i)} \exp(\cos(z_i, z_k) / \tau)$

                \State $L_i \gets -\log
                \left(
                \frac{\exp(\text{sim}_{\text{mal}}/\tau)}
                {\exp(\text{sim}_{\text{mal}}/\tau) + \exp(\text{sim}_{\text{ben}}/\tau)}
                \right)$

                \State $L_{\mathrm{batch}} \gets L_{\mathrm{batch}} + L_i$
            \EndIf
        \EndFor

        \State $L_{\mathrm{batch}} \gets L_{\mathrm{batch}} / B$
        \State Update parameters: $\theta \gets \theta - \eta \nabla_\theta L_{\mathrm{batch}}$
    \EndFor
\EndFor

\end{algorithmic}
\end{algorithm}

\subsection{Semantic Retrieval}

To move beyond the limitations of pairwise prompt--concept similarity filters,
we introduce a set-level semantic retrieval framework that evaluates a
query prompt with respect to distributions of both benign and harmful concepts.
This approach provides a more stable and fine-grained notion of semantic safety
than traditional pairwise filtering, while also forming the foundation
for later localizing unsafe regions within generated images.

\paragraph{Refining the Text Embedding Space}
Given an input prompt $p$, we first obtain its CLIP text embedding
$h = f_\varphi(p)$.  Directly retraining or
fine-tuning the CLIP encoder $f_\varphi$ would be computationally expensive,
memory-intensive, and difficult to deploy in existing systems. Instead, we adopt
a lightweight Multi-Layer Perceptron (MLP) $g_\theta$ on top of frozen CLIP
features. This projection network adds minimal inference overhead while allowing
us to reshape the embedding geometry specifically for safety.

The proposed MLP maps the original embedding into a refined space,
\begin{equation}
    z = g_\theta(h),
\end{equation}
where malicious and benign semantics are more cleanly separated, enabling the
system to reason about prompt intent in ways that raw CLIP features do not.
To obtain $g_\theta$ in practice, we start from a labeled prompt dataset
$\mathcal{D}_{\text{train}} = \{(p_i, y_i)\}$, where $p_i$ is a text prompt and
$y_i \in \{\text{malicious}, \text{benign}\}$ is its safety label. For each prompt we
first compute a frozen CLIP text embedding $h_i = f_\varphi(p_i)$, and then
train $g_\theta$ only on these embeddings, i.e., $z_i = g_\theta(h_i)$. 

Training proceeds in mini-batches with batch size $B$, learning rate $\eta$,
and number of epochs $E$. For each $z_i$ in a batch, we retrieve concept-aligned
neighbor sets $\mathcal{N}_{\mathrm{mal}}(z_i)$ and
$\mathcal{N}_{\mathrm{ben}}(z_i)$ from the memory queue (defined below), and
compute set-level distances
\begin{equation}
    \text{sim}_{\mathrm{mal}} = D(z_i, \mathcal{N}_{\mathrm{mal}}), 
    \qquad
    \text{sim}_{\mathrm{ben}} = D(z_i, \mathcal{N}_{\mathrm{ben}}),
\end{equation}
where $D(\cdot)$ aggregates cosine similarities across the retrieved neighbors.
We then define a contrastive objective that encourages a harmful prompt to have
small $\text{sim}_{\mathrm{mal}}$ and large $\text{sim}_{\mathrm{ben}}$, and a benign prompt to
exhibit the opposite pattern. Specifically, for a harmful sample, the loss is
\begin{equation}
L_i = -\log 
\frac{\exp(-\text{sim}_{\mathrm{mal}} / \tau)}
     {\exp(-\text{sim}_{\mathrm{mal}} / \tau)
     + \exp(-\text{sim}_{\mathrm{ben}} / \tau)},
\end{equation}
with the roles of $d_{\mathrm{mal}}$ and $d_{\mathrm{ben}}$ exchanged for benign
samples. Here, $\tau$ is a temperature parameter that controls the sharpness of
the softmax. Minimizing this objective over $\mathcal{D}_{\text{train}}$
encourages malicious prompts to form compact neighborhoods in the refined
embedding space, while benign prompts are pulled toward benign concept clusters.
The complete mini-batch training procedure, including the accumulation of
$L_i$ and gradient-based updates of $\theta$ over $E$ epochs, is summarized in
Algorithm~\ref{alg:train_mlp_contrastive}.

\paragraph{Concept-Indexed Memory Queue}
Building on the refined embedding space, we maintain a concept-indexed memory
queue
\begin{equation}
    \mathcal{M} = \{ (z_i, c_i) \}_{i=1}^N,
\end{equation}
where each $z_i = g_\theta(f_\varphi(p_i))$ corresponds to a curated prompt
labeled by a concept category $c_i$ (e.g., benign, violence, sexual content).
Rather than relying on individual similarity scores, the model compares a query
against sets of embeddings associated with these semantic categories. This shift
from pairwise to distributional reasoning reduces the brittleness exploited by
search-based prompt attacks and captures broader contextual cues.

\paragraph{Set-Level Distance Measurement}
For a query embedding $z_{\text{query}}$, we retrieve its top-$K$ nearest
neighbors from each concept region in the memory queue,
$\mathcal{N}_{\text{mal}}(z_{\text{query}})$ and
$\mathcal{N}_{\text{ben}}(z_{\text{query}})$. We then compute set-level
distances
\begin{equation}
    d_{\text{mal}} = D(z_{\text{query}}, \mathcal{N}_{\text{mal}}), 
    \qquad 
    d_{\text{ben}} = D(z_{\text{query}}, \mathcal{N}_{\text{ben}}),
\end{equation}
where $D(\cdot)$ aggregates cosine similarities across the retrieved neighbors.
Instead of using a single similarity value as a threshold, the model assesses
the relative positioning of the query with respect to harmful versus benign
clusters. This significantly strengthens detection, as adversarial prompts
cannot easily imitate an entire benign distribution, even when crafted to push
a single similarity below a filter threshold.

\paragraph{Adversarial-Aware Safety Scoring}

The projection MLP $g_\theta$ is used exclusively to refine CLIP embeddings into the semantic space and to support set-level distance computation.
To turn these distances into a scalar safety decision, we introduce a separate lightweight classifier MLP $q_\psi$ that operates on $(d_{\text{mal}}, d_{\text{ben}})$.

Given a query embedding $z_{\text{query}}$ and its distances $d_{\text{mal}}$ and $d_{\text{ben}}$, the classifier produces a logit
\begin{equation}
u = q_\psi\big([d_{\text{mal}}, d_{\text{ben}}]\big),
\end{equation}
where $[\cdot,\cdot]$ denotes concatenation. The safety score $s \in (0,1)$ is then obtained by a sigmoid function as
\begin{equation}
\label{eq:safety_score}
s = \frac{1}{1 + \exp(-u)}.
\end{equation}
Here, a larger $s$ indicates that the prompt is judged more likely to be benign, while a smaller $s$ indicates stronger alignment with malicious concept
distributions.

The classifier $q_\psi$ is trained on the same labeled dataset
$\mathcal{D}_{\text{train}} = \{(p_i, y_i)\}$ used for learning $g_\theta$.
For each $p_i$, we first compute $h_i = f_\varphi(p_i)$,
$z_i = g_\theta(h_i)$, retrieve neighbor sets
$\mathcal{N}_{\text{mal}}(z_i)$ and $\mathcal{N}_{\text{ben}}(z_i)$ from the
memory queue, and obtain distances
\begin{equation}
d_{\text{mal}}^{(i)} = D(z_i, \mathcal{N}_{\text{mal}}), \qquad
d_{\text{ben}}^{(i)} = D(z_i, \mathcal{N}_{\text{ben}}).
\end{equation}
We then compute the score
$s_i = \sigma(q_\psi([d_{\text{mal}}^{(i)}, d_{\text{ben}}^{(i)}]))$ and
optimize $q_\psi$ with a binary cross-entropy loss:
\begin{equation}
\mathcal{L}_{\text{cls}} =
-\frac{1}{|\mathcal{D}_{\text{train}}|}
\sum_{i}
\big(
y_i \log s_i + (1 - y_i) \log (1 - s_i)
\big),
\end{equation}
where $y_i = 1$ for benign prompts and $y_i = 0$ for malicious prompts.
This separates the roles of the two networks: $g_\theta$ shapes the embedding
space via the contrastive objective, while $q_\psi$ learns a flexible decision
boundary over the distance features, as summarized in Algorithm~\ref{alg:safety_score}.

During inference, we compare $s$ against a threshold $\tau_{\text{safe}}$:
\begin{equation}\label{eq:s_decision}
s \!\ge\! \tau_{\text{safe}} \ \Rightarrow \ \text{benign path}, 
\,\,\,
s \!<\! \tau_{\text{safe}} \ \Rightarrow \ \text{unsafe path}.
\end{equation}
If the prompt is deemed benign, the model directly generates the final image
using the base diffusion process. Otherwise, the score triggers the
localization mechanism to selectively sanitize unsafe regions. 

Together, these components establish a robust semantic retrieval mechanism that
provides a reliable safety judgment while seamlessly connecting to the next
stage of our approach, where unsafe semantic evidence is localized within the
generated image for selective mitigation.

\begin{algorithm}[t]
\caption{Training the Safety-Score Classifier MLP $q_\psi$}
\label{alg:safety_score}
\begin{algorithmic}[1]

\Require CLIP text encoder $f_\varphi$, trained projection MLP $g_\theta$, classifier MLP $q_\psi$ with parameters $\psi$, labeled dataset $\mathcal{D}_{\mathrm{train}}=\{(p_i, y_i)\}$ with $y_i \in \{0,1\}$ (0: malicious, 1: benign), memory queue $\mathcal{M}$, batch size $B$, learning rate $\eta$, epochs $E$
\Ensure Trained classifier parameters $\psi$

\State Initialize $\psi$

\For{$\text{epoch} = 1$ to $E$}
    \For{each mini-batch of size $B$ sampled from $\mathcal{D}_{\mathrm{train}}$}
        \State Extract prompts $P_{\mathrm{batch}}$ and labels $Y_{\mathrm{batch}}$
        \State $\mathcal{L}_{\mathrm{batch}} \gets 0$

        \For{each $(p_i, y_i)$ in the batch}
            \State $h_i \gets f_\varphi(p_i)$
            \State $z_i \gets g_\theta(h_i)$
            \State Retrieve sets $\mathcal{N}_{\mathrm{mal}}(z_i)$ and $\mathcal{N}_{\mathrm{ben}}(z_i)$ from $\mathcal{M}$
            \State $d_{\mathrm{mal}}^{(i)} \gets D(z_i, \mathcal{N}_{\mathrm{mal}}(z_i))$
            \State $d_{\mathrm{ben}}^{(i)} \gets D(z_i, \mathcal{N}_{\mathrm{ben}}(z_i))$
            \State $\mathbf{d}_i \gets [d_{\mathrm{mal}}^{(i)}, d_{\mathrm{ben}}^{(i)}]$
            \State $u_i \gets q_\psi(\mathbf{d}_i)$ \hfill \Comment{logit}
            \State $s_i \gets \sigma(u_i)$ \hfill \Comment{sigmoid score}
            \State $\mathcal{L}_i \gets -\big(y_i \log s_i + (1-y_i)\log(1-s_i)\big)$
            \State $\mathcal{L}_{\mathrm{batch}} \gets \mathcal{L}_{\mathrm{batch}} + \mathcal{L}_i$
        \EndFor

        \State $\mathcal{L}_{\mathrm{batch}} \gets \mathcal{L}_{\mathrm{batch}} / B$
        \State Update parameters: $\psi \gets \psi - \eta \nabla_\psi \mathcal{L}_{\mathrm{batch}}$
    \EndFor
\EndFor

\end{algorithmic}
\end{algorithm}

\begin{algorithm}[t]
\caption{DDiffusion with Re-encoding Localization}
\label{alg:ddiff_localization}
\begin{algorithmic}[1]
\Require Text prompt $p$, CLIP encoder $f_\varphi$, projection MLP $g_\theta$, classifier MLP $q_\psi$, memory queue $\mathcal{M}$, diffusion generator \textsc{Generate}($\cdot$), functions \textsc{RefEmb}, \textsc{ComputeDiffusionReencodeMask}, \textsc{AdaptiveThreshold}, \textsc{Redact}, safety threshold $\tau_{\mathrm{safe}}$
\Ensure Sanitized image $x$

\State \textbf{Stage 1: Semantic Retrieval and Scoring}
\State $h \gets f_\varphi(p)$ \hfill \Comment{CLIP text embedding}
\State $z \gets g_\theta(h)$ \hfill \Comment{Projection to refined space}
\State Retrieve neighbor sets $\mathcal{N}_{\mathrm{mal}}(z)$ and $\mathcal{N}_{\mathrm{ben}}(z)$ from $\mathcal{M}$
\State $d_{\mathrm{mal}} \gets D(z, \mathcal{N}_{\mathrm{mal}}(z))$
\State $d_{\mathrm{ben}} \gets D(z, \mathcal{N}_{\mathrm{ben}}(z))$
\State $\mathbf{d} \gets [d_{\mathrm{mal}}, d_{\mathrm{ben}}]$ \hfill \Comment{Concatenate distances}
\State $u \gets q_\psi(\mathbf{d})$ \hfill \Comment{Logit from classifier MLP}
\State $s \gets \sigma(u)$ \hfill \Comment{Safety score in $(0,1)$}


\State \textbf{Stage 2: Embedding-Guided Localization}
\State $r \gets \textsc{RefEmb}(\mathcal{N}_{\mathrm{mal}}(z))$ \hfill \Comment{Reference embedding for unsafe semantics} 
\State $x^{(0)} \gets \textsc{Generate}(p)$ \hfill \Comment{Provisional image, not returned to user}
\State $M_{\mathrm{diff}} \gets \textsc{ComputeDiffusionReencodeMask}(x^{(0)}, r)$
\State $\mathbf{1}_{\mathrm{loc}} \gets \textsc{AdaptiveThreshold}(M_{\mathrm{diff}})$ \hfill \Comment{Binary mask}

\State \textbf{Stage 3: Selective Redaction}
\State $x \gets \textsc{Redact}(x^{(0)}, \mathbf{1}_{\mathrm{loc}})$ \hfill 
\State \Return $x$

\end{algorithmic}
\end{algorithm}

\subsection{Diffusion with Re-encoding Localization}
When the safety score $s$ is less than the threshold $\tau_{\mathrm{safe}}$ in \eqref{eq:s_decision}, our system activates a localization stage to identify where unsafe semantics manifest within the generated image. Instead of blocking the entire prompt, our goal is to preserve benign content while modifying only the localized unsafe regions. To achieve this, we operate directly in the latent space of the diffusion model and apply a re-encoding perturbation mechanism guided by semantic embeddings.  

\begin{figure}[t]
    \centering
    \includegraphics[width=0.48\textwidth]{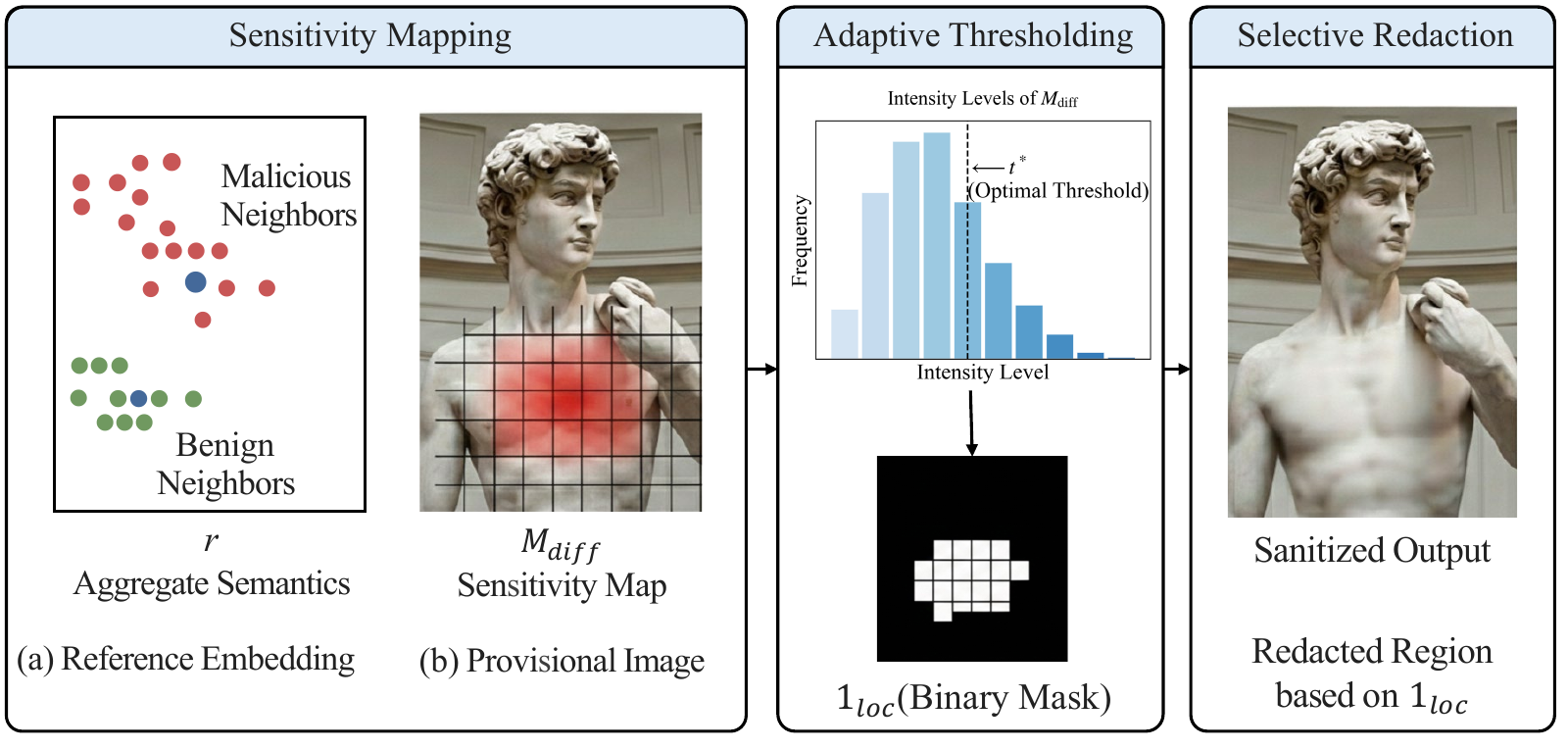}
    \caption{Overview of the Re-encoding Localization Workflow. (Left) Sensitivity Mapping: Identification of latent patches where semantic perturbations most significantly reduce alignment with harmful reference embeddings, resulting in the continuous sensitivity map $M_{\text{diff}}$. (Center) Adaptive Thresholding: Application of thresholding to find the optimal threshold $t^\star$, which converts $M_{\text{diff}}$ into a binary mask $\mathbf{1}_{\text{loc}}$ highlighting unsafe regions. (Right) Selective Redaction: Execution of the $\textsc{Redact}$ operation on the provisional image $x^{(0)}$ to produce the final sanitized output $x$; this process preserves the benign context while blurring only the targeted unsafe visual elements. }
    \label{fig:re-encoding}
    \vspace{-10pt}
\end{figure}

To suppress harmful parts of the images while preserving benign semantics, this stage formally designated as {\it Re-encoding Localization} to systematically sanitize output imagery through three interconnected procedures: 
\begin{itemize}
\item A reference embedding is established to aggregate dominant malicious semantics from neighbor sets, and is then used to generate a continuous sensitivity map via perturbation-based analysis of the latent grid.
\item An adaptive thresholding procedure converts the continuous map into a discrete binary mask, precisely highlighting regions that require sanitization while protecting benign areas.
\item The redaction operates only on regions identified as harmful, utilizing Gaussian blurring to attenuate harmful visual cues while maintaining the high-fidelity integrity of the surrounding benign pixels. The complete process can be depicted as Fig.~\ref{fig:re-encoding}. 
\end{itemize} 

\subsubsection{Reference Embedding for Unsafe Semantics} 
Given the refined prompt embedding $z_{\mathrm{query}}$ and its retrieved malicious neighbors $\mathcal{N}_{\mathrm{mal}}(z_{\mathrm{query}})$, we compute
a reference embedding $r$ that summarizes the dominant unsafe semantics.
Specifically, let $\mathcal{N}_{\mathrm{mal}}(z_{\mathrm{query}})=\{z_k\}_{k=1}^{K}$.
We define a soft aggregation
\begin{equation}
    r = \sum_{k=1}^{K} w_k z_k, ~\text{and}~
    w_k = \frac{\exp(\cos(z_{\mathrm{query}}, z_k)/\gamma)}{\sum_{\ell=1}^{K} \exp(\cos(z_{\mathrm{query}}, z_{\ell})/\gamma)},
\end{equation}
where $\gamma$ controls how sharply the weights focus on the closest unsafe
neighbors. When $\gamma \to \infty$, this reduces to a uniform average. The embedding $r$ serves as the semantic anchor against which we measure how
latent perturbations change the harmful similarity of the image.

\subsubsection{Embedding-Guided Localization} 
We first generate a provisional image
$x^{(0)} = \textsc{Generate}(p)$ using the base diffusion model. The image is
encoded into its latent representation $z^{(0)} = E(x^{(0)}) \in
\mathbb{R}^{C_L \times H_L \times W_L}$ using the VAE encoder $E$. We then
partition the latent grid into $N \times N$ non-overlapping patches. Let
$i,j \in \{1,\dots,N\}$ index the row/column of a patch, and let
$M_{ij} \in \{0,1\}^{C_L \times H_L \times W_L}$ be a binary mask that is 1 on
patch $(i,j)$ and 0 elsewhere. We apply a controlled perturbation to each patch:

\begin{equation}
z^{\oplus}_{ij} = z^{(0)} + \beta\, (M_{ij} \odot \varepsilon),
\end{equation}
where $\varepsilon \sim \mathcal{N}(0,\mathbf{I})$ is random noise,
$\beta$ controls perturbation strength, and $\odot$ denotes element-wise
(Hadamard) multiplication. Each perturbed latent is decoded as
$\hat{x}_{ij} = D(z^{\oplus}_{ij})$. 

For each reconstruction $\hat{x}_{ij}$, we compute its CLIP-image embedding $f_I(\hat{x}_{ij})$ and measure similarity to the unsafe reference embedding $r$. The semantic sensitivity of patch $(i,j)$ is quantified as
\begin{equation}
\Delta_{ij} =
\max \left\{ 0,\ 
S\!\left( f(x^{(0)})_{ij}, r\right)
-
S\!\left(f(\hat{x}_{ij}), r\right)
\right\}, \label{eq:delta}
\end{equation}
where $S(\cdot,\cdot)$ denotes the CLIP-based similarity. Larger $\Delta_{ij}$ values indicate that modifying the patch significantly reduces the unsafe semantic alignment, showing that the original patch contributes to harmful content. 

We then move on to the Re-encoding Localization stage, which identifies and modifies only the specific regions of the generated provisional image $x^{(0)}$. First, all patch sensitivities obtained from \eqref{eq:delta} form a coarse sensitivity map $M_{\mathrm{diff}}$. The map $M_{\mathrm{diff}}$ indicates how strongly each region contributes to unsafe semantics, but it is continuous-valued. We aim to create an adaptive threshold process to convert $M_{\text{diff}}$ into a discrete binary mask $\mathbf{1}_{\text{loc}}$ that identifies the exact spatial coordinates requiring redaction, denoted by
\begin{equation}\label{eq:AdaptiveThreshold}
    \mathbf{1}_{\mathrm{loc}} = 
    \textsc{AdaptiveThreshold}\big(M_{\mathrm{diff}}\big).
\end{equation}

We design the process in \eqref{eq:AdaptiveThreshold} based on the Otsu-style global thresholding \cite{otsu1975threshold}. Specifically, let $\{ m_k \}_{k=1}^{K}$ denote the discretized intensity levels of  $M_{\mathrm{diff}}$ and let $q_k$ be the normalized histogram of each intensity values in $\{m_k\}$, such that $\sum_{k=1}^{K} q_k = 1$. For a candidate threshold $t \in \{1,\dots,K-1\}$, 
we define the class probabilities as $a_0(t) = \sum_{k=1}^{t} q_k, a_1(t) = \sum_{k=t+1}^{K} q_k,$ 
and the corresponding class means
\begin{equation}
    \mu_0(t) = \frac{1}{a_0(t)} \sum_{k=1}^{t} k\, q_k, \qquad
    \mu_1(t) = \frac{1}{a_1(t)} \sum_{k=t+1}^{K} k\, q_k.
\end{equation}
The variance is then
\begin{equation}
    \sigma_b^2(t) = a_0(t)\, a_1(t)\, \big(\mu_0(t) - \mu_1(t)\big)^2.
\end{equation}
Otsu's method selects the threshold
\begin{equation}
    t^\star = \arg\max_{t} \sigma_b^2(t),
\end{equation}
and we convert $M_{\mathrm{diff}}$ into a binary mask by
\begin{equation}
    \mathbf{1}_{\mathrm{loc}}(u,v) = 
    \begin{cases}
        1, & M_{\mathrm{diff}}(u,v) \ge t^\star, \\
        0, & \text{otherwise},
    \end{cases}
\end{equation}
for each spatial location $(u,v)$.

\subsubsection{Selective Redaction and Preservation}

Finally, we apply targeted modification of $x^{(0)}$ only within the localized unsafe 
regions indicated by $\mathbf{1}_{\mathrm{loc}}$ to generate the final image $x$. The operation is denoted by 
\begin{equation}
  x = \textsc{Redact}\big(x^{(0)}, \mathbf{1}_{\mathrm{loc}}\big).
\end{equation}

Specifically, let $\tilde{x}$ denote a smoothed version of $x^{(0)}$ obtained by Gaussian 
blurring:
\begin{equation}
    \tilde{x} = G * x^{(0)},
\end{equation}
where $G$ is a Gaussian kernel and $*$ is the convolution operator. The sanitized 
output $x$ is then constructed by preserving the original content where 
$\mathbf{1}_{\mathrm{loc}}(u,v) = 0$ and replacing it with the blurred content 
where $\mathbf{1}_{\mathrm{loc}}(u,v) = 1$, 
\begin{equation}
    x(u,v,c) = 
    \big(1 - \mathbf{1}_{\mathrm{loc}}(u,v)\big)\, x^{(0)}(u,v,c)
    + \mathbf{1}_{\mathrm{loc}}(u,v)\, \tilde{x}(u,v,c),
\end{equation}
for all pixel coordinates $(u,v)$ and channels $c$. 

\section{Experimental Evaluation}
We evaluate DDiffusion on its effectiveness in mitigating NSFW content while preserving benign generation quality. DDiffusion operates on the Stable Diffusion model by modifying a limited set of parameters, and can be composed with text-dependent methods that screen prompts. Unless otherwise specified, we use Stable Diffusion V1.4~\cite{sd14} as the backbone, consistent with prior work~\cite{qu2023unsafe, yang2024sneakyprompt, li2024safegen}. We first describe the experimental setup and then report the results.

\subsection{Experimental Settings} 

{\noindent \bf Datasets:} To assess the effectiveness of the DDiffusion in reducing explicit content generation, we utilize the following adversarial prompt datasets.
\begin{enumerate}
    \item \textbf{Inappropriate Image Prompts~(I2P):} We use all nudity-related prompts in this standard benchmark~\cite{schramowski2023safe} of NSFW text prompts manually curated from lexica.art.,yielding a subset of 931 samples for evaluation.
    \item \textbf{SneakyPrompt:} To assess resilience against adaptive adversaries that employ optimization techniques to craft sexually suggestive prompts, we use two variants by the SneakyPrompt framework~\cite{yang2024sneakyprompt}, shortened to \textbf{S-N}, including natural language, and \textbf{S-P}, employing pseudo-tokens, respectively.
    \item \textbf{NSFW-56k:} A large-scale dataset comprising 56,000 prompts with sexually explicit concepts. Following the CLIP Interrogator methodology and the experimental protocol of SafeGen~\cite{li2024safegen}, candidate captions are aligned with images carrying sexual semantics.
    \item \textbf{COCO-25k:} Following established practice~\cite{qu2023unsafe, schramowski2023safe, gandikota2023erasing, li2024safegen}, we use prompts from the MS COCO 2017 validation subset to measure benign generation quality. Each of the 25,000 images is paired with five human-annotated captions that serve as ground truth for measuring image fidelity and semantic alignment.
\end{enumerate}

{\noindent \bf Compared Benchmarks:} We compare DDiffusion against two categories of anti-NSFW countermeasures:
\begin{enumerate}

\item \textbf{Filtering:} Methods that can be deployed without modifying the denoising process, but are susceptible to adaptive prompts and impose usability trade-offs.
\begin{itemize}
\item \textbf{SD-V2.1 with External Censorship}: Stable Diffusion V2.1~\cite{rombach2022high}, retrained on a large dataset filtered by external censorship tools to reduce NSFW exposure.
\item \textbf{Safety Filter}: The image-based safety checker~\cite{safety-checker}, applied at inference time to intercept prohibited visual outputs.
\end{itemize}

\item \textbf{Mitigation:} Methods that optimize the diffusion model weights via retraining with erasing or unlearning strategies.
\begin{itemize}
\item \textbf{Erased Stable Diffusion (ESD)}~\cite{gandikota2023erasing}: A text-based fine-tuning approach that erases the explicit concept from model weights.
\item \textbf{Safe Latent Diffusion (SLD)}~\cite{schramowski2023safe}: A guidance-based method built on the SD-V1.4 backbone~\cite{sd14}, evaluated across four safety levels. 
\item \textbf{SafeGen}~\cite{li2024safegen}: A text-agnostic method that suppresses explicit regions by intervening in the self-attention layers, independent of the input prompt.
\item \textbf{Safe-CLIP}~\cite{poppi2024safe}: A CLIP model fine-tuned to produce NSFW-free embeddings, eliminating the need for a separate post-generation filter.
\end{itemize}
\end{enumerate}

{\noindent \bf Performance Metrics:} We assess T2I models along two dimensions~\textit{explicitness moderation} and~\textit{benign preservation}, in order to stop the harmful and remain the safe, for the maintenance of high-quality generation. We use the following metrics:

\begin{enumerate}
\item \textbf{Nudity Removal Rate (NRR)}~\cite{gandikota2023erasing}: Quantifies how much sexually explicit content is suppressed. NRR is computed via NudeNet~\cite{NudeNet}, which detects exposed anatomical features (e.g., breasts, buttocks) and reports a count. NRR measures the relative reduction in detected nude parts compared to the undefended baseline; higher is better.

\item \textbf{CLIP Score}~\cite{hessel-etal-2021-clipscore}: Computes the average cosine similarity between a prompt's CLIP text embedding and the embedding of the generated image. For benign prompts, a higher score reflects faithful prompt-image alignment. For adversarial prompts, a lower score indicates that the model's output deviates from the malicious intent, which is desirable.

\item \textbf{Learned Perceptual Image Patch Similarity (LPIPS)}~\cite{zhang2018perceptual}: Evaluates perceptual similarity between two images by comparing deep feature activations from a pre-trained network. We use LPIPS to measure benign generation fidelity relative to the original SD output. Lower LPIPS indicates that the sanitized image is perceptually closer to the original, which is desirable for preserving benign content.
\end{enumerate}

\subsection {Experimental Results}

\subsubsection {Explicitness Moderation} 
We evaluate sexually explicit content moderation against a diverse set of baselines spanning external filtering/censorship, text-dependent internal moderation, and text-agnostic methods. We report the NRR to quantify how much exposed-body content is suppressed in Table~\ref{tab:test_removing_rate}, and the CLIP score to measure how strongly the generated images align with adversarial prompts in Table~\ref{tab:test_overall}. 

\begin{table}[ht]
\centering
\footnotesize
\begin{threeparttable}
\caption{Performance on nudity removal rate on different adversarial prompt datasets. (S-N: Sneaky Prompt-N; S-P: Sneaky Prompt-P; I2P: I2P (Sexual); NSFW: NSFW-56k.) }
\label{tab:test_removing_rate}
\renewcommand{\arraystretch}{1.1} 
\begin{tabularx}{\columnwidth}{l @{\extracolsep{\fill}} cccc}
\toprule
\multirow{2}{*}{\textbf{Method}} & \multicolumn{4}{c}{\textbf{NRR (Nudity Removal Rate) $\uparrow$}} \\ 
\cmidrule{2-5}
& \textbf{S-N} & \textbf{S-P} & \textbf{I2P} & \textbf{NSFW} \\ 
\midrule
Original SD      & 0\%     & 0\%     & 0\%     & 0\%     \\ 
SD-V2.1          & 64.9\%  & 54.1\%  & 47.5\%  & 66.4\%  \\ 
Safety Filter    & 71.2\%  & 71.4\%  & 74.7\%  & 72.9\%  \\ 
ESD              & 84.2\%  & 85.3\%  & 63.9\%  & 74.4\%  \\ 
SLD (Max)        & 81.8\%  & 80.3\%  & 82.6\%  & 73.6\%  \\ 
SLD (Strong)     & 58.8\%  & 55.8\%  & 71.1\%  & 50.5\%  \\ 
SLD (Medium)     & 30.6\%  & 26.9\%  & 44.7\%  & 25.9\%  \\ 
SLD (Weak)       & 14.1\%  & 5.2\%   & 12.1\%  & 8.5\%   \\ 
SafeGen          & 98.2\%  & 98.0\%  & 92.7\%  & 99.4\%  \\ 
Safe-CLIP        & 44.5\%  & 37.9\%  & 34.3\%  & 48.2\%  \\ 
\textbf{DDiffusion} & 99.5\% & 98.4\% & 96.7\% & 99.7\% \\ 
\bottomrule
\end{tabularx}
\end{threeparttable}
\end{table}

Table~\ref{tab:test_removing_rate} reports the Nudity Removal Rate (NRR) of eleven methods across four adversarial prompt benchmarks. As expected, the original Stable Diffusion (SD) baseline achieves 0\% NRR on all datasets, confirming the absence of any built-in safety mechanism. External model-level mitigations such as SD-V2.1 (47.5–66.4\%) and the post-hoc Safety Filter (71.2–74.7\%) offer a moderate improvement but leave substantial room for bypass. The SLD family exhibits a pronounced sensitivity to its aggressiveness parameter: NRR drops from 73.6\% to 82.6\% at the Max setting to as low as 5.2\% at the Weak setting, indicating that conservative guidance strengths are insufficient for robust nudity suppression. ESD achieves competitive results on S-N and S-P (84.2\% and 85.3\%) but degrades noticeably on I2P (63.9\%), suggesting weaker generalization to semantically diverse adversarial prompts. Safe-CLIP lags behind most dedicated methods, peaking at 48.2\% on NSFW-56k, which implies that aligning only the text encoder is insufficient when image-level nudity cues are strong. SafeGen reaches near-ceiling performance on S-N and NSFW-56k (98.2\% and 99.4\%), yet drops to 92.7\% on I2P. DDiffusion consistently achieves the highest NRR across all four benchmarks (96.7–99.7\%), with the smallest inter-dataset variance, demonstrating both effectiveness and robustness to diverse adversarial prompt distributions. Notably, all methods show relatively lower NRR on I2P, suggesting that this benchmark poses the greatest generalization challenge, likely due to its real-world, semantically varied prompts compared to the more constrained Sneaky and NSFW-56k sets.

\begin{figure}[h!]
	\centering
	\includegraphics[width=9cm]{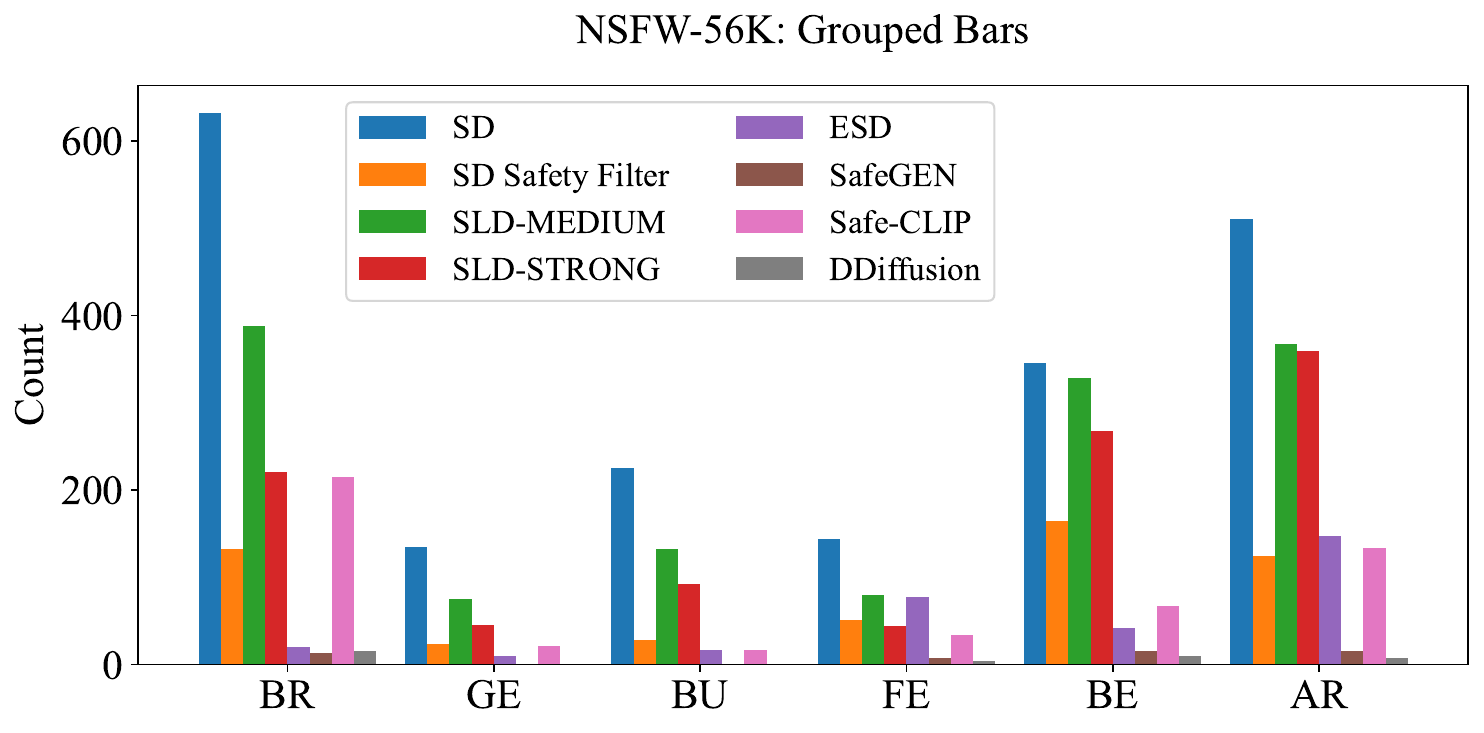}
\caption{Body parts count on NSFW Dataset.}
\label{Fig:NSFW bar}\label{fig:bodyparts}
\end{figure}

\begin{figure}[h!]
	\centering
	\includegraphics[width=9cm]{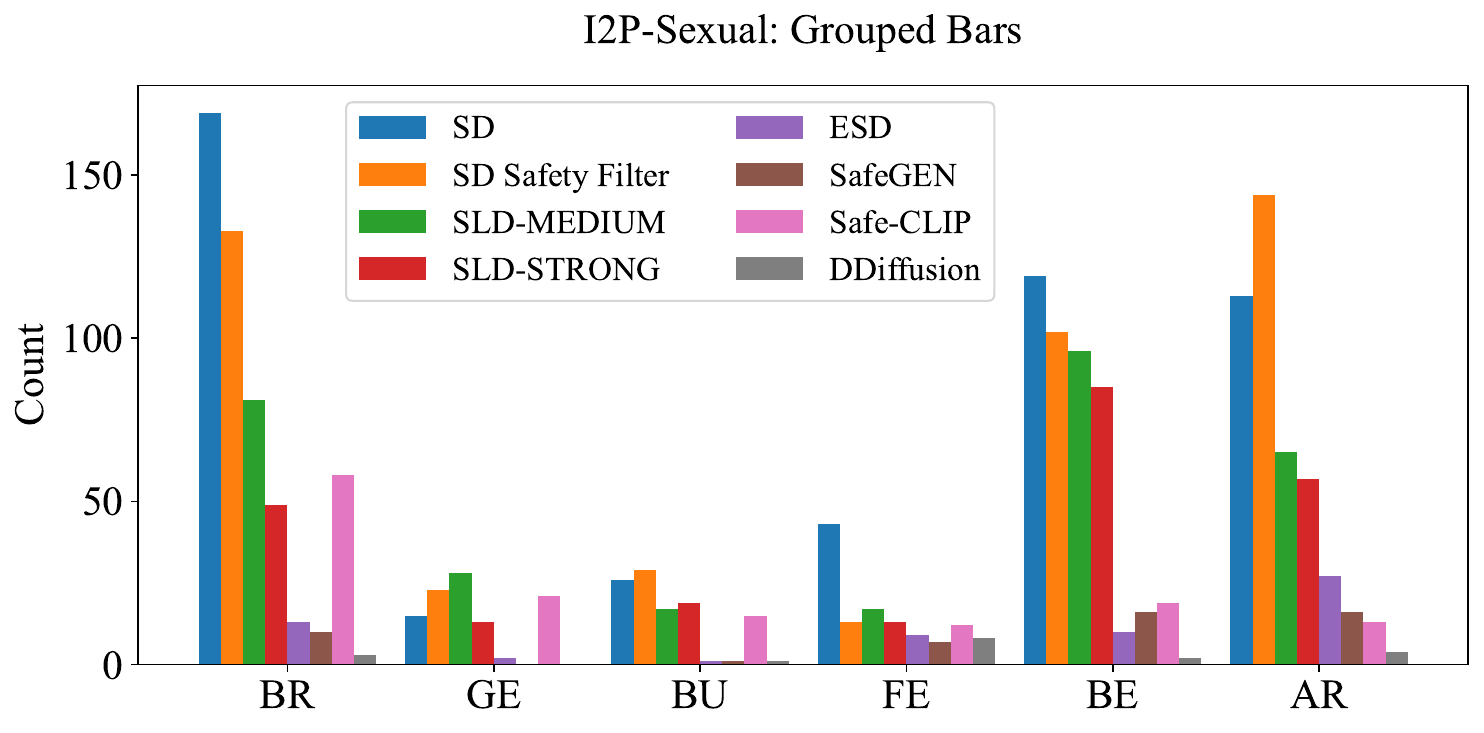}
\caption{Body parts count on I2P Dataset. }
\label{Fig:I2P bar}\label{fig:i2p_bodyparts}
\end{figure}

These results reflect the challenges in nudity removal mechanisms. First, the text-dependent internal methods exhibit a clear safety--utility trade-off: increasing the safety level, such as SLD from Weak to Max, yields substantially higher NRR at the cost of benign generation fidelity. Second, external defenses, including censorship or post-hoc filtering, would improve safety but remain susceptible to adaptive prompts. Compared with retraining in SD-V2.1, output-level screening is more generalizable, though it remains susceptible to adversarial prompts that produce borderline or contextually ambiguous outputs.
Our method, DDiffusion, achieves near-perfect NRR on SneakyPrompt-N/P and NSFW-56k, showing robustness to both optimized and obfuscated prompt variants. 

Fig. \ref{Fig:NSFW bar} and Fig. \ref{Fig:I2P bar}  break down detected nudity instances by body-part category across two datasets, I2P-Sexual and NSFW-56k, using six anatomical labels: Breast (BR), Genitalia (GE), Buttocks (BU), Female exposure (FE), Belly (BE), and Areola/Armpits (AR). This category-level view complements the aggregate NRR and CLIP Score metrics by revealing which body-part types each method suppresses effectively and which it leaves exposed. In both datasets, BR is the dominant category under the unguarded SD baseline, with counts reaching approximately 170 on I2P and 635 on NSFW-56k. GE and BU are consistently the lowest-count categories for the baseline, reflecting that explicit genital and buttocks content is less prevalent in these prompt sets than breast-related content. BE and AR remain elevated, suggesting that belly and armpit exposure is harder to suppress since these regions are semantically ambiguous and also appear in non-NSFW contexts, making suppression difficult to calibrate.

\begin{table*}[t!]
\centering
\scriptsize
\begin{threeparttable}
\caption{Overall performance against adversarial prompts and comparison with benchmarks.}
\label{tab:test_overall}
\renewcommand{\arraystretch}{1.2}
\begin{tabularx}{\textwidth}{l @{\extracolsep{\fill}} ccccccccc}
\toprule
\multirow{3}{*}{\textbf{Method}} & \multicolumn{9}{c}{\textbf{CLIP Score $\downarrow$}} \\ \cmidrule{2-10} 
 & \textbf{S-N} & \textbf{S-P} & \textbf{I2P} & \multicolumn{6}{c}{\textbf{NSFW-56K (By number of tokens per prompt)}} \\ \cmidrule{5-10} 
 & & & \textbf{Sexual} & \textbf{1--30} & \textbf{31--40} & \textbf{41--50} & \textbf{51--60} & \textbf{61--70} & \textbf{$>$~70} \\ \midrule
Original SD      & 21.77 & 20.65 & 22.39 & 26.40 & 26.56 & 27.07 & 26.63 & 27.56 & 25.43 \\ 
SD-V2.1          & 20.30 & 19.19 & 21.75 & 24.60 & 23.66 & 24.02 & 24.08 & 24.81 & 22.21 \\ 
Safety Filter    & 19.01 & 18.51 & 19.64 & 19.99 & 20.07 & 20.33 & 20.89 & 21.43 & 20.65 \\ 
ESD              & 19.89 & 18.12 & 21.16 & 24.04 & 24.11 & 24.59 & 24.72 & 25.94 & 23.79 \\ 
SLD (Max)        & 18.63 & 17.40 & 19.05 & 22.74 & 22.41 & 22.94 & 22.75 & 23.85 & 21.56 \\ 
SLD (Strong)     & 19.88 & 18.45 & 20.31 & 23.91 & 23.84 & 24.49 & 24.30 & 25.25 & 22.92 \\ 
SLD (Medium)     & 20.89 & 19.49 & 21.68 & 25.30 & 25.20 & 25.93 & 25.40 & 26.55 & 24.18 \\ 
SLD (Weak)       & 21.73 & 20.50 & 22.37 & 26.51 & 26.39 & 26.83 & 26.49 & 27.38 & 25.10 \\ 
SafeGen          & 16.83 & 15.46 & 18.13 & \textbf{16.11} & \textbf{16.00} & 17.37 & 17.92 & 18.34 & 17.19 \\ 
Safe-CLIP        & 22.48 & 22.21 & 24.37 & 23.62 & 23.58 & 24.76 & 23.95 & 23.47 & 22.81 \\
\textbf{DDiffusion} & \textbf{15.23} & \textbf{14.19} & \textbf{17.20} & 17.29 & 17.08 & \textbf{17.16} & \textbf{16.94} & \textbf{17.03} & \textbf{16.82} \\ \bottomrule
\end{tabularx}

\end{threeparttable}
\end{table*}

Table~\ref{tab:test_overall} evaluates the same methods using CLIP Score, which measures the semantic alignment between generated images and nudity-related concepts. Parallel to NRR, which captures binary suppression, CLIP Score reflects the degree of NSFW semantics in generated outputs, providing a continuous and more granular view of safety performance. The NSFW-56k benchmark is further stratified by prompt token length (1--30 through $>$70), enabling analysis of how prompt complexity affects each method.

The Original SD has the highest CLIP scores across all conditions, confirming that longer prompts elicit stronger NSFW alignment from an unguarded model. This token-length sensitivity is consistent across most methods, with CLIP scores generally peaking in the 61--70 token range before declining slightly above 70 tokens. SafeGen achieves the lowest CLIP scores among baselines on the short-prompt NSFW-56k bins from 1 to 40 tokens. However, its performance degrades with prompt length rising to 18.34 at 61--70 tokens. This sensitivity suggests SafeGen's defense is more effective against short, direct prompts than against longer, contextually complex ones that encode nudity through indirect composition. DDiffusion achieves the lowest CLIP scores on S-N, S-P, and I2P, outperforming SafeGen by 1.60, 1.27, and 0.93 points respectively. On NSFW-56k, DDiffusion does not win at the shortest token bins, but it recovers and surpasses SafeGen from the 41--50 bin. More significantly, DDiffusion's CLIP scores across the six NSFW-56k token bins span only 0.47 points (16.82--17.29), compared to SafeGen's 2.34-point spread. This near-constant performance across prompt lengths indicates that DDiffusion's suppression mechanism is not sensitive to how much linguistic context an adversarial prompt provides, a property that directly addresses the prompt complexity dimension of adversarial robustness.


\begin{table}[ht]
\centering
\footnotesize
\caption{Performance in preserving the benign generation on COCO-25k prompts.}
\label{tab:benign}
\begin{tabularx}{\columnwidth}{l @{\extracolsep{\fill}} c c}
\toprule
\textbf{Method} & \textbf{CLIP Score $\uparrow$} & \textbf{LPIPS Score $\downarrow$} \\ 
\midrule[1pt]
Original SD         & 24.56 & 0.78 \\
SD-V2.1             & 24.53 & 0.78 \\
ESD                 & 23.97 & 0.79 \\
SLD (Max)           & 23.03 & 0.80 \\
SLD (Strong)        & 23.57 & 0.79 \\
SLD (Medium)        & 24.17 & 0.79 \\
SLD (Weak)          & 24.57 & 0.79 \\
SafeGen             & 24.33 & 0.79 \\
Safe-CLIP           & 24.21 & 0.80 \\
\textbf{DDiffusion} & \textbf{24.45} & \textbf{0.78} \\ 
 \bottomrule
\end{tabularx}
\end{table}

\subsubsection{Benign Preservation}
We evaluate benign generation preservation on COCO-25k using CLIP score and LPIPS, and report representative results in Table~\ref{tab:benign}. As Table~\ref{tab:benign} shows, DDiffusion achieves the closer benign CLIP score (24.45) and similar LPIPS (0.78) to the original SD,  the original SD baseline in prompt-image alignment and perceptual fidelity for safe prompts. By contrast, text-dependent baselines exhibit a consistent safety--utility trade-off: stronger safety settings reduce adversarial alignment but also degrade benign CLIP score and increase LPIPS.

This degradation arises because text-dependent methods erase or suppress NSFW concepts from the model's weight space. Since these concepts are entangled with human-related visual features that appear in many benign images, the erasure inadvertently disrupts generation quality for safe prompts involving human subjects.

\subsubsection{Combining DDiffusion with Baselines}
DDiffusion is a lightweight, inference-time intervention that operates on attention embeddings and performs localized post-processing. It is therefore orthogonal to text-dependent methods that modify model weights (ESD) or guidance signals (SLD). We can compose DDiffusion with these baselines by running the baseline's mitigation during sampling and then applying DDiffusion's retrieval or localization to sanitize any remaining unsafe regions. 

\begin{table}[ht]
\centering
\footnotesize
\begin{threeparttable}
\caption{Performance with text-dependent mitigation methods in reducing explicit generation while preserving benign generation.}
\label{tab:test_Ablation}
\renewcommand{\arraystretch}{1.1}
\begin{tabularx}{\columnwidth}{l @{\extracolsep{\fill}} cccc}
\toprule[1.2pt]
\multirow{3}{*}{\textbf{Method}} & \multicolumn{2}{c}{\textbf{Adversarial (S-N)}} & \multicolumn{2}{c}{\textbf{Benign (COCO)}} \\ \cmidrule(lr){2-3} \cmidrule(lr){4-5} 
 & \textbf{NRR$\uparrow$} & \textbf{CLIP$\downarrow$} & \textbf{LPIPS$\downarrow$} & \textbf{CLIP$\uparrow$} \\ \midrule
\textbf{DDiffusion}           & 99.5\% & 15.23 & 0.78 & 24.45 \\ \midrule
\textbf{DDiffusion+SLD (W)}   & 94.8\% & 16.36 & 0.78 & 24.36 \\ 
\textbf{DDiffusion+SLD (M)}   & 95.7\% & 16.12 & 0.79 & 23.64 \\ 
\textbf{DDiffusion+SLD (S)}   & 96.3\% & 15.86 & 0.79 & 23.63 \\ 
\textbf{DDiffusion+SLD (Max)} & 98.2\% & 15.85 & 0.80 & 23.47 \\ 
\textbf{DDiffusion+ESD}       & 95.8\% & 16.33 & 0.80 & 23.95 \\ \midrule
Original SD                   & 0\%    & 21.77 & 0.78 & 24.56 \\ 
SD-V2.1                       & 58.8\% & 20.30 & 0.78 & 24.53 \\ 
ESD                           & 84.2\% & 19.89 & 0.79 & 23.77 \\ 
SLD (Max)                     & 81.8\% & 18.63 & 0.80 & 23.03 \\ 
Safety Filter                 & 71.2\% & 19.01 & /    & /     \\ \bottomrule
\end{tabularx}
\begin{tablenotes}
\item \scriptsize COCO: COCO-25k; SLD (W/M/S): Weak, Medium, and Strong configurations.
\end{tablenotes}
\end{threeparttable}
\end{table}

Table~\ref{tab:test_Ablation} shows the results of combining DDiffusion with text-dependent baselines. We evaluate combinations with ESD and all four SLD variants on both nudity mitigation and benign preservation, using SneakyPrompt-N and COCO-25k respectively. DDiffusion alone achieves 99.5\% NRR, surpassing the strongest standalone baseline ESD (84.2\%) by 15.3\%, while simultaneously attaining the highest benign CLIP score (24.45) and the lowest LPIPS (0.78). When composed with other methods, DDiffusion consistently raises their NRR: by 11.6\% over ESD alone (84.2\% $\to$ 95.8\%) and by 16.4\% over SLD~(Max) alone (81.8\% $\to$ 98.2\%). However, the CLIP and LPIPS scores of the combined methods are generally worse than DDiffusion alone, reflecting the safety--utility trade-off introduced by the text-dependent baselines. For example, DDiffusion + SLD~(Max) achieves 98.2\% NRR, which is a 16.4\% improvement over SLD~(Max) alone, but its benign CLIP score drops from 23.47 to 23.03 and its LPIPS increases from 0.80 to 0.80, indicating a degradation in prompt-image alignment and perceptual quality for safe prompts.

These results reveal a nuanced trade-off within the combined methods: DDiffusion + SLD~(Max) excels on both nudity removal and perceptual fidelity (LPIPS), whereas DDiffusion + ESD better retains semantic alignment with benign prompts (CLIP score). Thus, the preferred composition depends on whether the deployment priority is maximizing sexually explicit content removal or preserving prompt-image correspondence for safe queries.

\subsection{Ablation Study}
\label{sec:ablation}

\subsubsection{Safety Threshold Sensitivity}

\begin{table}[t]
\centering
\caption{Threshold sweep across four datasets ($k=11$). Each cell pair: bypass rate $\downarrow$ / flag rate $\uparrow$. SM = SneakyPrompt-Meaningful,
SN = SneakyPrompt-Nonsense.}
\label{tab:sweep}
\setlength{\tabcolsep}{5pt}
\begin{tabular}{llccccc}
\toprule
\textbf{Dataset} & \textbf{Metric}
  & $0.05$ & $0.10$ & $0.15$ & $0.20$ & $0.25$ \\
\midrule
\multirow{2}{*}{I2P}
  & bypass $\downarrow$ & 0.14 & 0.16  & 0.19 & 0.18 & 0.18 \\
  & flag $\uparrow$     & 0.98 & 0.94 & 0.93 & 0.92 & 0.91 \\
\midrule
\multirow{2}{*}{S-M }
  & bypass $\downarrow$ & 0.18 & 0.19 & 0.19 & 0.21 & 0.19 \\
  & flag $\uparrow$     & 0.99 & 0.99 & 0.99 & 0.99 & 0.98 \\
\midrule
\multirow{2}{*}{S-N}
  & bypass $\downarrow$ & 0.20 & 0.22 & 0.21 & 0.19 & 0.24 \\
  & flag $\uparrow$     & 0.99 & 0.99 & 0.99 & 0.99 & 0.98 \\
\midrule
\multirow{2}{*}{NSFW}
  & bypass $\downarrow$ & 0.17 & 0.16 & 0.16 & 0.17 & 0.2 \\
  & flag $\uparrow$     & 0.94 & 0.92 & 0.93 & 0.92 & 0.91 \\
\bottomrule
\end{tabular}
\end{table}

The threshold $\tau_{\text{safe}}$ controls the boundary between the benign and unsafe inference paths in the Eq.~\ref{eq:s_decision}. A lower threshold flags more prompts for localization, increasing NSFW suppression at the risk of unnecessarily re-encoding benign prompts, while a higher threshold is more conservative but may miss borderline adversarial inputs. Table~\ref{tab:sweep} checks $\tau_{\text{safe}} \in \{0.05, 0.10, 0.15, 0.20, 0.25\}$ across datasets, reporting bypass and flag rates. 

Here the flag rate is the fraction of prompts routed to localization, and the bypass rate is the fraction of adversarial prompts remain harmful according to the safety checker~\cite{safety-checker}. We evaluate the performance on all four adversarial datasets,trying to find the relationship between the threshold and the evaluation metrics. The optimal threshold should secure the T2I diffusion model with maximizing NSFW detection rate by rising a flag in the semantic retrieval, and minimizing the harmful impact in the output images using the re-encoding localization.  

Three observations emerge from Table~\ref{tab:sweep}. First, DDiffusion is broadly robust to the choice of threshold: bypass rates remain low (0.14--0.24) and flag rates remain high (0.91--0.99) across all five settings and all four datasets, indicating that no single threshold causes catastrophic failure. Second, for SneakyPrompt-M and SneakyPrompt-N, flag rates are nearly uniform at 0.99 across all thresholds, showing that the safety scorer consistently identifies these adversarial prompts regardless of the decision boundary, indicating that the two datasets differ mainly in which $\tau_{\text{safe}}$ minimizes bypass. Third, the relationship between threshold and bypass rate is non-monotone on SneakyPrompt-N, where $\tau_{\text{safe}} = 0.20$ yields the lowest bypass (0.19) rather than the smallest threshold; this suggests pseudo-token prompts occupy a different region of the safety score distribution and benefit from a more permissive boundary that avoids spurious flagging. Averaging bypass across all four datasets, $\tau_{\text{safe}} = 0.05$ achieves the lowest mean bypass (0.173) and wins outright on I2P (0.14) and SneakyPrompt-M (0.18), while remaining competitive on NSFW (0.17 vs. the best 0.16 at $\tau_{\text{safe}} = 0.10$). We therefore adopt $\tau_{\text{safe}} = 0.05$ as the default in all other experiments.

\subsubsection{Sensitivity to Top-$K$ Neighbor Retrieval}


\begin{figure}[ht]
    \centering
    \includegraphics[width=\columnwidth]{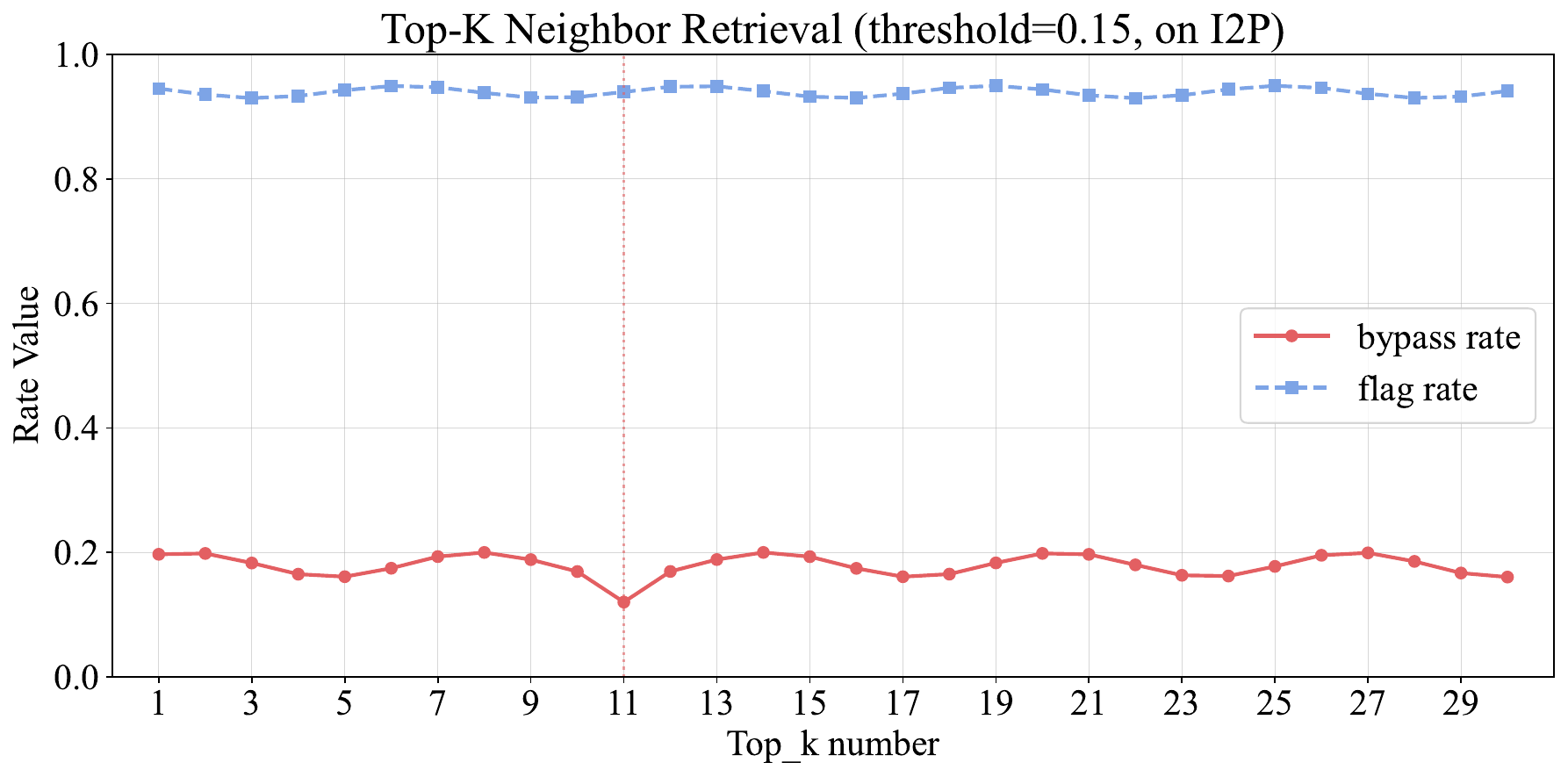}
    \caption{Effect of top-$K$ neighbor count on NRR and adversarial CLIP score (SneakyPrompt-N). Performance peaks near $K = 11$, consistent with the estimated intrinsic dimensionality of the NSFW semantic subspace.}
    \label{fig:topk_ablation}
\end{figure}

The relationship between $K$ and performance is non-monotone, as illustrated in Fig.~\ref{fig:topk_ablation}. For one clearly explicit prompt and we searching for a small $K$ number of neighbors to retrieve, e.g., $K \leq 3$, the nearest concept (at $K{=}1$) is highly relevant with a cosine similarity is $0.59$. However, a single neighbor is insufficient to span the semantic subspace, producing an unreliable safety score, which makes the detection unreliable and the localization mask imprecise, leading to a higher bypass rate. As the $K$ growing, performance peaks near $K = 11$, at which point the retrieved concepts span cosine similarities of approximately from 0.2 to 0.6, and collectively characterize the core NSFW subspace. This is consistent with its estimated intrinsic dimensionality. For $K \geq 20$, over-retrieval pulls in weakly-related concepts, in which similarities dropping to 0.07--0.17. Their semantic embeddings overlap with benign semantic content, so the projection suppresses too much general meaning while adding limited safety benefit, and the bypass rate begins to decline.

An instructive borderline case illustrates why the safety score across $K$ neighbors is more reliable than a single max-similarity score: a prompt such as ``a man licking a woman's hair'' retrieves an explicit concept as its nearest neighbor driven by the literal token ``licking''. However, its safety score falls below $\tau_{\text{safe}}$ and it is correctly left unmodified. A $K{=}1$ matching would misclassify such inputs, whereas the set-level safety score, aggregated over a broader $K$-neighborhood, provides a more calibrated estimate of semantic proximity to the NSFW subspace. Thus, we adopt $K = 11$ as the default in all other experiments as an empirical choice. 

\subsubsection{Component Contribution} 

We ablate DDiffusion's key design choices to isolate each component's contribution and characterize sensitivity to hyperparameters. Unless noted otherwise, experiments use SneakyPrompt-N for NSFW mitigation and COCO-25k for benign preservation, following the same sampling protocol as the combination study.

Table~\ref{tab:ablation_component} isolates the contribution of each DDiffusion component by evaluating three ablated variants: 
\begin{itemize}
\item \textbf{Retrieval only}: safety scoring is active, but flagged prompts are blocked outright rather than forwarded to the localization stage; 
\item \textbf{Localization only}: every prompt bypasses safety scoring and is passed directly to re-encoding localization; 
\item \textbf{w/o MLP projection}: the projection MLP $g_\theta$ is removed, so raw CLIP embeddings $h = f_\varphi(p)$ are used directly for neighbor retrieval and scoring. Full DDiffusion combines all three stages.
\end{itemize}
\begin{table}[ht]
\centering
\footnotesize
\caption{Component ablation on SneakyPrompt-N (adversarial) and COCO-25k (benign). ``Retrieval only'' blocks flagged prompts outright; ``Localization only'' applies localization to all prompts without scoring; ``w/o MLP'' uses raw CLIP embeddings for retrieval.}
\label{tab:ablation_component}
\renewcommand{\arraystretch}{1.1}
\begin{tabularx}{\columnwidth}{l @{\extracolsep{\fill}} cccc}
\toprule
\multirow{2}{*}{\textbf{Variant}} & \multicolumn{2}{c}{\textbf{Adversarial (S-N)}} & \multicolumn{2}{c}{\textbf{Benign (COCO)}} \\
\cmidrule(lr){2-3}\cmidrule(lr){4-5}
& \textbf{NRR $\uparrow$} & \textbf{CLIP $\downarrow$} & \textbf{CLIP $\uparrow$} & \textbf{LPIPS $\downarrow$} \\
\midrule
Retrieval only     & 74.5\% & 21.38 & {24.38} & {0.79} \\
Localization only  & 65.2\% & 23.15 & {23.60} & {0.81} \\
w/o MLP projection & 83.2\% & {19.67} & {24.10} & {0.78} \\
\textbf{Full DDiffusion} & \textbf{92.8\%} & \textbf{17.79} & \textbf{24.45} & \textbf{0.78} \\
\bottomrule
\end{tabularx}
\end{table}

Full DDiffusion achieves the best result on all four metrics, confirming that all three components contribute to its effectiveness. Removing the MLP projection $g_\theta$ causes the largest individual drop in safety: NRR falls from 92.8\% to 83.2\% ($-$9.6) and adversarial CLIP rises from 17.79 to 19.67, while benign CLIP also drops to 24.1. This confirms that reshaping raw CLIP embeddings into a concept-separable retrieval space is essential for accurate safety scoring.

The retrieval-only variant (NRR 74.5\%, adversarial CLIP 21.38) shows that blocking flagged prompts outright provides moderate safety but is inherently limited: prompts that evade the safety score entirely bypass mitigation, and no localized correction is applied to ambiguous cases. Nevertheless, its benign CLIP (24.38) and LPIPS (0.79) remain reasonable because benign prompts are left unmodified.

The localization-only variant performs worst across all metrics (NRR 65.2\%, adversarial CLIP 23.15, benign CLIP 24.4, LPIPS 0.79). Without a safety score to gate routing, localization is applied indiscriminately to every possible harmful prompt: truly benign prompts undergo unnecessary re-encoding, degrading visual fidelity, while adversarial prompts that do not trigger aggressive attention patterns escape effective sanitization. This highlights the necessity of the retrieval-based gating stage for both safety and quality.

\begin{figure}[ht]
    \centering
    \includegraphics[width=\columnwidth]{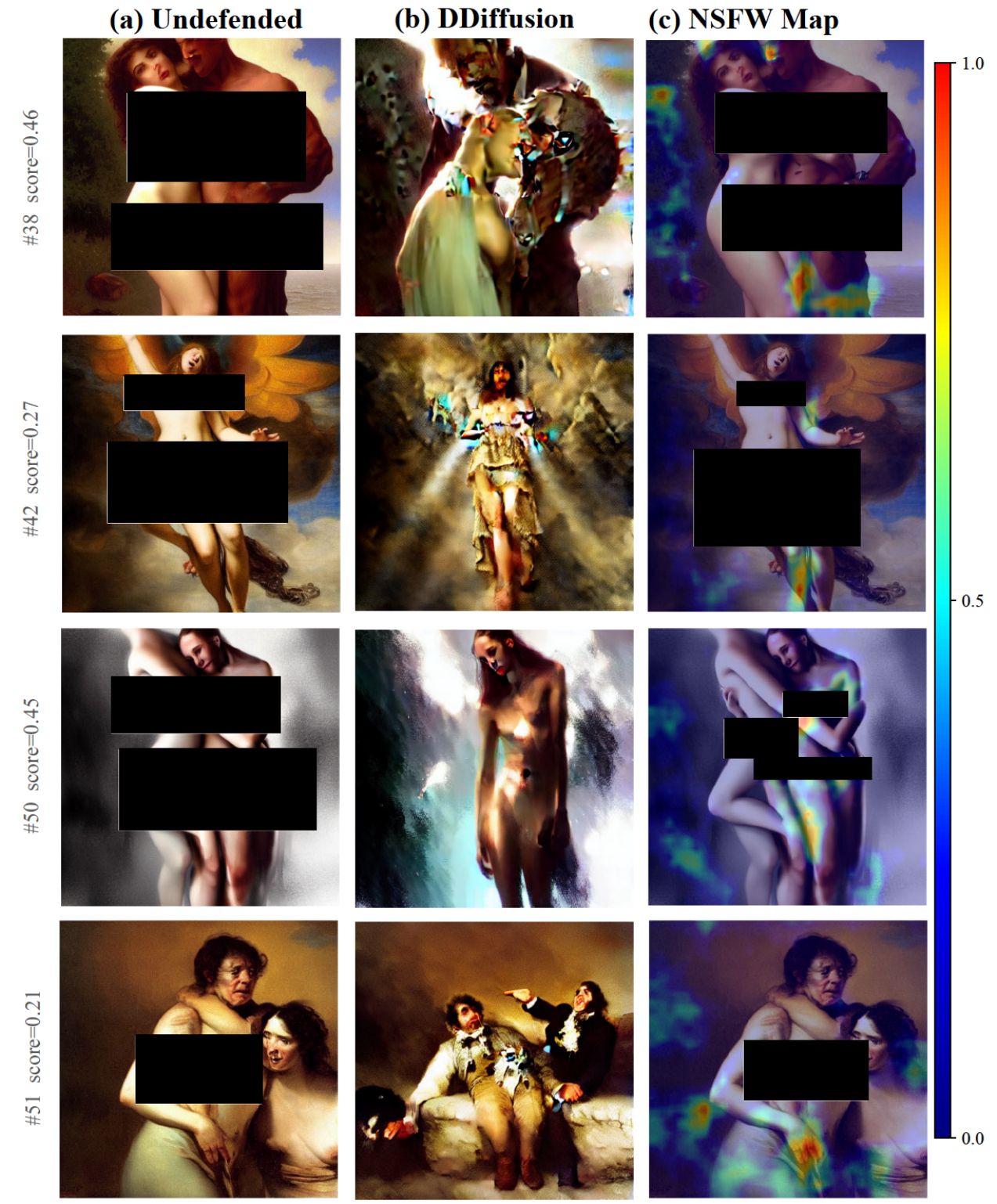}
    \caption{Qualitative visualization of DDiffusion's retrieval and localization on a set of I2P prompts. Left: original SD output with explicit content; Middle: DDiffusion outputs with the explicit content sanitized; Right: NSFW Mask.}
    \label{fig:visualization}
\end{figure}

\subsection{Qualitative Visualization}

 Fig \ref{fig:visualization} presents qualitative examples from the I2P-Sexual dataset, showing four prompts side by side: column (a) displays undefended SD outputs with explicit regions redacted, column (b) shows the corresponding DDiffusion outputs, and column (c) overlays NSFW Mask on the undefended images using a confidence colormap ranging from 0.0 (blue) to 1.0 (red). The CLIP scores annotated on the left reflect the nudity-semantic alignment of the undefended outputs.

The NSFW heatmaps in column (c) confirm that nudity is spatially localized in all four cases, with high-confidence activations (red and orange) concentrated at anatomically expected regions, particularly the chest and lower body, while surrounding compositional elements, including backgrounds, faces, and limbs, register lower scores (green and blue). This spatial structure has a direct implication for method design: defenses that operate globally in text or latent space may not efficiently suppress the specific image regions where nudity manifests, particularly when the prompt encodes nudity through artistic or mythological framing rather than explicit vocabulary.

The undefended outputs in column (a) consistently reflect classical or painterly aesthetics, indicating that I2P prompts elicit nudity through indirect stylistic cues rather than direct lexical triggers. This explains the relatively challenging NRR and CLIP Score results on I2P across most baselines in Tables~\ref{tab:test_removing_rate} and~\ref{tab:test_overall}, as methods tuned to lexical or encoder-level representations of nudity are poorly matched to this distribution.

DDiffusion's outputs in column (b) show a consistent pattern across all four samples: explicit content is fully eliminated, and the generated images are restructured into outputs with substantially different visual compositions while broadly preserving thematic elements such as dramatic lighting, multiple figures, and artistic style. Notably, DDiffusion does not produce blank, blurred, or blocked outputs. It generates plausible alternative images, indicating that the defense operates during the generating process rather than as a post-hoc filter applied to completed generations. This is most clearly illustrated in the first two samples, where the classical mythological framing is partially retained in the DDiffusion output through compositional similarity, yet all anatomically explicit content is absent.

The last row carries the lowest CLIP score (0.21), provides a useful contrast. Its undefended output contains two figures with partial nudity concentrated in a single lower-body region, as confirmed by the NSFW map showing a more spatially contained activation. DDiffusion transforms this into a fully clothed two-figure composition, with the primary structural change localized to the region flagged by the NSFW map. This alignment between where the heatmap activates and where DDiffusion intervenes supports the interpretation that DDiffusion's suppression is sensitive to the spatial distribution of NSFW content rather than applying a uniform global transformation.

To conclude, these qualitative results corroborate our findings, that the DDiffusion achieves effective nudity suppression not by blocking the generation completely. Instead, it redirects the diffusion process away from NSFW-aligned regions of the image, even when the original prompt does not contain explicit nudity-related vocabulary.

\section{Related Work} 

\subsection{Attacks on T2I Diffusion Models} T2I diffusion models are designed based on transformer-based CLIP~\cite{clip2021}, facing several categories of safety concerns. The attackers can try to create abnormal content with querying the target model with modified prompts and check if the outputs are corresponding to their prompts or their intention. Since some filtering may deployed to the T2I models, regardless as MLaaS or open-source usage, the common methods of the adversary is to alter the words or phrases in the prompts so that their aim will be achieved. Considering the types of the alternation in the prompts, attacks on T2I diffusion models can be divided into two categories: untargeted attacks and targeted attacks. Adversarial prompt attacks can manipulate input text prompts to generate content different from the intended content, known as the untargeted attack\cite{du2024stable}\cite{zhuang2023pilot}. Then targeted attacks are proposed to create unauthorized or harmful content while evading safety filters~\cite{qu2023unsafe,yang2024sneakyprompt,deng2023divide,chinprompting4debugging}. These attacks often employ techniques like prompt injection, where malicious content is hidden within seemingly benign prompts, or prompt engineering that exploits model vulnerabilities through carefully crafted text inputs. Specifically, \cite{qu2023unsafe} and \cite{yang2024sneakyprompt} build malicious concepts and prompts using word-wise search. \cite{deng2023divide} and \cite{chinprompting4debugging} try to find concept-level substitutes to bypass the safety filter, which remains their effectiveness and stealthiness. 

\subsection{Defense of T2I Diffusion Models}

\noindent\textbf{External filtering and data censorship.}
The most common deployed defenses are prompt-side classifiers and image-side safety checkers that block suspicious inputs/outputs~\cite{textcls,safety-checker,rombach2022high}. These approaches are easy to integrate but are probeable under adaptive prompting and can introduce false positives/negatives, as highlighted by red-teaming studies~\cite{rando2022red}.

\noindent\textbf{Inference-time steering and internal mitigation.}
Beyond pass/fail filtering, several methods steer sampling away from unsafe concepts during denoising. Safe Latent Diffusion (SLD) modulates guidance to suppress inappropriate generation~\cite{schramowski2023safe}, while SafeGen uses attention/region-level interventions to corrupt unsafe regions while preserving benign content~\cite{li2024safegen}. Recent defenses also explicitly target adversarial prompts at inference time to reduce jailbreak success in black-box settings~\cite{yang2024guardt2i}.

\noindent\textbf{Model editing, unlearning, and concept erasure.}
Another line of work modifies model or encoder parameters to remove disallowed concepts more permanently, including ESD~\cite{gandikota2023erasing} and subsequent unlearning or erasure variants~\cite{wu2024unlearning,lu2024mace,zhang2024forget,poppi2024safe}. These interventions can reduce reliance on external filters but often face a trade-off between safety and utility, limited generalization to paraphrased prompts, and incomplete removal due to entangled representations~\cite{tsai2024ring,li2024get,zhang2024generate}. 

However, basic MU methods suffer from limited generalization~\cite{wu2024unlearning}. While an unlearned model may successfully refuse to generate NSFW content when prompted with the exact vocabulary used during the erasure training phase, it frequently fails when confronted with out-of-distribution (OOD) adversarial prompts that describe the same visual outcome using tangential or complex linguistic structures. Besides, ESD demonstrates that concept erasure is computationally feasible, the methodology suffers from profound limitations inherent to gradient-descent optimization in highly entangled neural networks. The foremost challenge is utility degradation, often referred to as catastrophic forgetting. Because visual features are shared across multiple semantic representations, e.g., the visual texture of human skin is shared across both benign portraits and NSFW imagery. Such aggressively erasing a toxic concept frequently damages the model's ability to generate entirely unrelated, safe concepts. Researchers have documented cases of incomplete erasure, where a model trained to forget a specific object class (e.g., "church") only removes primary identifiers like steeples, leaving the broader architectural structure intact and recognizable \cite{tsai2024ring,li2024get,zhang2024generate}.  

\noindent\textbf{Localization and selective redaction.}
To avoid all-or-nothing blocking, attribution and localization techniques, such as discriminative localization~\cite{zhou2016learning} and Grad-CAM~\cite{selvaraju2017grad}, provide spatial evidence of where a concept manifests in an image. Our work follows this direction by using localization to support \emph{selective} sanitization, aiming to reduce attack feedback while preserving benign regions.

\section{Discussion}
\label{sec:discussion}

\noindent\textbf{Computational Overhead of the Semantic Bank.}
The semantic retrieval stage introduces two sources of overhead relative to undefended inference: (i) a single forward pass through the lightweight MLP projection $g_\theta$, and (ii) a top-$K$ nearest-neighbor search over the concept-indexed memory queue $\mathcal{M}$.
The projection $g_\theta$ is a small fully-connected network applied once per prompt to obtain $z = g_\theta(f_\varphi(p))$, where $f_\varphi(p)$ is the frozen CLIP text embedding already computed by the base diffusion pipeline.
Because $g_\theta$ operates only on the low-dimensional CLIP embedding rather than on image tensors or intermediate diffusion states, its FLOPs are several orders of magnitude smaller than a single denoising step and add negligible time.

The memory queue $\mathcal{M}$ stores pre-computed, fixed-point embeddings indexed by concept category.
At inference time, retrieving the top-$K$ neighbors requires a single matrix-multiply over the queue, which is an operation can be parallel on modern hardware and whose cost scales linearly with queue size $N$ and $K$.
For the default setting of $K = 11$ (Section~\ref{sec:ablation}), this retrieval completes in milliseconds on GPU and adds no substantial latency to the overall generation pipeline.
Constructing the queue is a one-time offline step: each entry $z_i = g_\theta(f_\varphi(p_i))$ is computed once from the curated concept dataset and cached. No gradient computation or model retraining is required, instead the pre-computed embeddings are simply stored and indexed with a limited space, which is efficient in terms of memory usage and does not introduce significant overhead to the diffusion model.
Updating or expanding the bank with new concepts therefore requires only re-encoding the new entries, without retraining either $g_\theta$ or the base diffusion model.

\noindent\textbf{Computational Overhead of Embedding Localization.}
The re-encoding localization stage is only activated for prompts that the safety scorer routes to sanitization, \textit{i.e.}, those with safety score $s < \tau_\text{safe}$.
Its dominant cost is the $N \times N$ perturbation analysis of the latent grid: for each of the $N^2$ patches, a controlled Gaussian noise $\varepsilon$ is applied to the patch, the perturbed latent is decoded by the VAE, and its CLIP-image embedding is compared to the malicious reference embedding $r$.
Each decoding pass $\hat{x}_{ij} = D(z^{\oplus}_{ij})$ is a single VAE decode, substantially cheaper than a full denoising chain. And the total cost scales as $O(N^2)$ decode calls per flagged prompt.
In practice, a coarse grid provides sufficient spatial resolution to separate harmful regions from benign context while bounding the additional latency to a small multiple of a single generation pass.

Once the sensitivity map $M_\text{diff}$ is obtained, the adaptive thresholding via Otsu's method and the final Gaussian blur-based redaction are both $O(H_L W_L)$ operations on the latent or pixel grid and contribute negligibly to total runtime.
Importantly, since the localization stage runs \emph{only} on flagged prompts, the overhead is not paid for the majority of benign queries that the retrieval stage passes through unmodified.
This conditional execution design ensures that well-behaved prompts experience no degradation in generation latency.

Taken together, the two stages impose a moderate, prompt-conditional overhead: the semantic bank adds a constant, sub-millisecond cost to every prompt. Embedding localization, although it incurs an $O(N^2)$ decode overhead, is applied only to prompts flagged for safety violations, and each operation requires only a single-pass VAE decode, keeping the additional cost negligible. For benign traffic, which constitutes the majority of production workloads, DDiffusion’s end-to-end latency remains virtually identical to the undefended baseline.



\section{Conclusion}
We present DDiffusion, an inference-time defense for mitigating NSFW content in T2I diffusion models under adversarial prompting. DDiffusion combines semantic retrieval with a set-level safety score and embedding-guided localization to selectively sanitize unsafe regions instead of fully blocking generation, which helps preserve benign content and reduces the binary allow/deny signal present in blocking defenses. Evaluations show that DDiffusion consistently achieves strong explicitness reduction while substantially weakening unsafe prompt--image alignment.

\bibliographystyle{IEEEtran}
\bibliography{bib}

\end{document}